\documentclass[runningheads]{llncs}
\usepackage{tikz}
\usepackage{comment}
\usepackage{amsmath,amssymb} 
\usepackage{color}
\usepackage{adjustbox}
\usepackage{booktabs}

\usepackage[accsupp]{axessibility}  

\usepackage{graphicx}
\usepackage{amsmath}
\usepackage{amssymb}
\usepackage{booktabs}
\usepackage[utf8]{inputenc} 
\usepackage[T1]{fontenc}    
\usepackage{url}            
\usepackage{booktabs}       
\usepackage{amsfonts}       
\usepackage{nicefrac}       
\usepackage{microtype}      
\usepackage{xcolor}         
\usepackage{algorithm}
\usepackage{algorithmic}
\usepackage{subcaption}
\usepackage{enumitem}
\usepackage{pifont}
\usepackage{wrapfig}
\usepackage{diagbox}

\usepackage{bm}

\newcommand{\xmark}{\ding{55}}
\newcommand{\quotes}[1]{``#1''}

\DeclareMathOperator{\bb}{\mathbf{b}}
\DeclareMathOperator{\ab}{\mathbf{a}}
\DeclareMathOperator{\yb}{\mathbf{y}}
\DeclareMathOperator{\eb}{\mathbf{e}}
\DeclareMathOperator{\pb}{\mathbf{p}}
\DeclareMathOperator{\xb}{\mathbf{x}}
\DeclareMathOperator{\wb}{\mathbf{w}}
\DeclareMathOperator{\Ub}{\mathbf{U}}

\newcommand{\argmin}{\mathop{\rm argmin}}
\newcommand{\thetab}{{\bm{\theta}}}
\newcommand{\mub}{{\bm{\mu}}}

\usepackage[capitalize]{cleveref}
\crefname{section}{Sec.}{Secs.}
\Crefname{section}{Section}{Sections}
\Crefname{table}{Table}{Tables}
\crefname{table}{Tab.}{Tabs.}

\begin{document}
\pagestyle{headings}
\mainmatter
\def\ECCVSubNumber{3645}  

\title{Meta-Learning with Less Forgetting on  Large-Scale Non-Stationary Task Distributions} 

\titlerunning{Meta Learning Less Forgetting}
%
\author{Zhenyi Wang\inst{1}\index{van Author, First E.} \and
Li Shen\inst{2}$^{*}$\and
Le Fang\inst{1}  \and
Qiuling Suo\inst{1} \and
Donglin Zhan\inst{3} \and
Tiehang Duan\inst{4} \and
Mingchen Gao\inst{1}$^{*}$
}

\renewcommand{\thefootnote}{\fnsymbol{footnote}}
\footnotetext[1]{Co-correspondence author.}

\authorrunning{Z. Wang et al.}
%
\institute{State University of New York at Buffalo, USA \email{\{zhenyiwa, lefang, qiulings, mgao8\}@buffalo.edu} \and
JD Explore Academy, Beijing, China
\email{mathshenli@gmail.com}\\
\and
Columbia University, New York, NY, USA  \email{dz2478@columbia.edu }\\
\and
Meta, Seattle, WA, USA \email{tiehang.duan@gmail.com}\\}


\maketitle

\begin{abstract}

The paradigm of machine intelligence moves from purely supervised learning to a more practical scenario when many loosely related unlabeled data are available and labeled data is scarce. Most existing algorithms assume that the underlying task distribution is stationary. Here we consider a more realistic and challenging setting in that task distributions evolve over time. We name this problem as \textbf{S}emi-supervised meta-learning with \textbf{E}volving \textbf{T}ask di\textbf{S}tributions, abbreviated as \textbf{SETS}. Two key challenges arise in this more realistic setting: (i) how to use unlabeled data in the presence of a large amount of unlabeled out-of-distribution (OOD) data; and (ii) how to prevent catastrophic forgetting on previously learned task distributions due to the task distribution shift. We propose an \textbf{O}OD \textbf{R}obust and knowle\textbf{D}ge pres\textbf{E}rved semi-supe\textbf{R}vised meta-learning approach (\textbf{ORDER}) \footnote{we use ORDER to denote the task distributions sequentially arrive with some ORDER}, to tackle these two major challenges. Specifically, our ORDER introduces a novel mutual information regularization to robustify the model with unlabeled OOD data and adopts an optimal transport regularization to remember previously learned knowledge in feature space. In addition, we test our method on a very challenging dataset: SETS on large-scale non-stationary semi-supervised task distributions consisting of (at least) 72K tasks. With extensive experiments, we demonstrate the proposed ORDER alleviates forgetting on evolving task distributions and is more robust to OOD data than related strong baselines.

\end{abstract}

\section{Introduction}

Meta-learning \cite{metanet} focuses on learning lots of related tasks to acquire common knowledge adaptable to new unseen tasks, and has achieved great success in many computer vision problems \cite{guo2020learning,Wang_2019_ICCV,guo2021metacorrection}. Recently, it has been shown that additional unlabeled data is beneficial for improving the meta-learning model performance \cite{ren2018metalearning,li2019learning,liu2019learning} especially when labeled data is scarce. 
One common assumption of these works \cite{ren2018metalearning,li2019learning,liu2019learning} is that the task distribution (i.e., a family of tasks) is stationary during meta training. However, real-world scenarios are more complex and often require learning across different environments.

One instance is in personalized self-driving systems \cite{bae2020selfdriving}. 
Each task refers to learning a personal object recognition subsystem for an individual user. The labeled data (annotated objects) for each user is scarce and expensive, but unlabeled data (e.g., images captured by their in-car cameras) is abundant.
Learning useful information from these unlabeled data would be very helpful for learning the system. Moreover, the task distributions are not stationary, as users' driving behaviors are different across regions. Assume the learning system is first deployed in Rochester and later extended to New York. The original dataset for Rochester may be no longer available when training on New York users due to storage or business privacy constraints.  Similar scenarios occur when Amazon extends the market for the personalized conversation AI product, Alexa, to different countries \cite{CL2020,huang-etal-2020-semi}.

As can be seen from the above examples, the first significant challenge is that task distributions may evolve during the learning process; the learning system could easily forget the previously learned knowledge. Second, despite that the additional unlabeled data are potentially beneficial to improve model performance \cite{ren2018metalearning,li2019learning,liu2019learning}, blindly incorporating them may bring negative effects. This is due to the fact that unlabeled data could contain both useful (In-Distribution (ID)) and harmful (Out-Of-Distribution (OOD)) data. How to leverage such mixed (ID and OOD) unlabeled data in meta-learning is underexplored in previous literature.

In general, these challenges can be summarized into a new realistic problem setting that the model learns on a sequence of distinct task distributions without prior information on task distribution shift. We term this problem setup as \textit{semi-supervised meta-learning on evolving task distributions} (\textbf{SETS}). 
To simulate the task distribution shift in SETS, we construct a dataset by composing 6 datasets, including \textit{CIFARFS} \cite{bertinetto2019metalearning}, \textit{AIRCRAFT} \cite{maji2013finegrained}, \textit{CUB} \cite{WelinderEtal2010}, \textit{Miniimagenet} \cite{matching16}, \textit{Butterfly} \cite{chen2018finegrained} and \textit{Plantae} \cite{vanhorn2018inaturalist}. A model will be sequentially trained on the datasets in a meta-learning fashion, simulating the real scenarios of evolving environment.
 However, at the end of training on the last domain, the meta learned model could easily \textit{forget} the learned knowledge to solve \textit{unseen} tasks sampled from the task distributions trained previously.

In this paper, we aim to tackle the SETS problem, which requires the meta-learning model to adapt to a new task distribution and retain the knowledge learned on previous task distributions. To better use unlabeled data of each task, we propose a regularization term driven by mutual information, which measures the mutual dependency between two variables. We first propose an OOD detection method to separate the unlabeled data for each task into ID and OOD data, respectively. Then, we minimize the mutual information between class prototypes and OOD data to reduce the effect of OOD on the class prototypes. At the same time, we maximize the mutual information between class prototypes and ID data to maximize the information transfer. We derive tractable sample-based variational bounds specialized for SETS for mutual information maximization and minimization for practical computation since the closed-form of probability distribution in mutual information is not available. 

To mitigate catastrophic forgetting during the learning process, we propose a novel and simple idea to reduce forgetting in feature space.
Specifically, in addition to storing the raw data as used in standard continual learning, we also store the data features of memory tasks. 
To achieve this goal, we use optimal transport (OT) to compute the distance between the feature distribution of memory tasks at the current iteration and the feature distribution of memory tasks stored in the memory. This ensures minimal feature distribution shift during the learning process. The proposed OT framework can naturally incorporate the unlabeled data features of previous tasks to further improve the memorization of previous knowledge. Besides, as the feature size is much smaller than the raw data, this makes it scalable to large-scale problems with a very long task sequence.

Our contributions are summarized as three-fold:
\begin{itemize}[noitemsep,nolistsep,leftmargin=*]
    \item  To our best knowledge, this is the first work considering the practical and challenging \textit{semi-supervised meta-learning on evolving task distributions} (SETS) setting. 
    \item  We propose a novel algorithm for SETS setting to exploit the unlabeled data and a mutual information inspired regularization to tackle the OOD problem and an OT regularization to mitigate forgetting.
    \item Extensive experiments on the constructed dataset with at least 72K tasks demonstrate the effectiveness of our method compared to strong baselines.
\end{itemize}

 \section{Related Work}\label{sec:related_work} 

\textbf{Meta Learning} \cite{metanet} focuses on rapidly adapting to new unseen tasks by learning prior knowledge through training on a large number of similar tasks. Various approaches have been proposed \cite{matching16,protonet17,Edwards2017TowardsAN,finn17a,PMAML2018,pmlr-v48-santoro16,antoniou2018train,munkhdalai2017meta,learn2learn2016,SNAILICLR18,lee2019meta,Wang2020Bayesian,metakernel,metafusion}. All of these methods work on the simplified setting where the model is meta trained on a stationary task distribution in a fully supervised fashion, which is completely different from our proposed SETS.
On the other hand, online meta-learning \cite{finn2019online} aims to achieve better performance on future tasks and stores all the previous tasks to avoid forgetting with a small number of tasks. Furthermore, \cite{wang2021meta,Wang_2022_CVPR} has proposed a meta learning method to solve large-scale task stream with sequential domain shift. However, our method is different from the previous work in several aspects. Specifically, (i) we additionally focus on leveraging unlabeled data and improving model robustness; (ii) we only store a small number of tasks to enable our method to work in our large-scale setting; (iii) we consider that the task distributions could be evolving and OOD-perturbed during meta training under the semi-supervised learning scenario.  In summary, SETS is more general and realistic than online meta-learning and other settings.

\textbf{Continual Learning (CL)} aims to maintain previous knowledge when learning on a sequence of tasks with distribution shift. Many works mitigate catastrophic forgetting during the learning process
\cite{lopezpaz2017gradient,AGEM19,riemer2018learning,yoon2018lifelong,EWC16,nguyen2017variational,ebrahimi2020adversarial,aljundi2018memory,IL2MCV,aljundi2019taskfree,pmlr-v162-wang22v}. 
Semi-supervised continual learning (SSCL) \cite{lee2019overcoming,wang2021ordisco,smith2021memoryefficient} is a recent extension of CL to the semi-supervised learning setting. SETS is completely different from CL and SSCL in several  aspects: (i) our SETS focuses on a large-scale task sequence (more than 72K), while CL and SSCL are not scalable to this large-scale setting; (ii) our goal is to enable the model to generalize to the \textit{unseen} tasks from all the previous task distributions, while SSCL works on the generalization to the testing data of previous tasks.

\textbf{Semi-supervised Few-shot Learning (SSFSL)} \cite{ren2018metalearning,li2019learning,liu2019learning} aims to generalize to unseen semi-supervised tasks when meta training on semi-supervised few-shot learning tasks in an \textit{offline} learning manner. It assumes that the few-shot learning task distributions are stationary during meta training. Unlike these works, we work on a more challenging and realistic setting of evolving task distributions in an \textit{online} learning manner. Meanwhile, SETS also aims at mitigating catastrophic forgetting for the previous learned tasks during online learning process as well, which makes it more challenging and practical to real-world scenarios.

\begin{wraptable}{r}{0.65\textwidth}
\captionsetup{width=1.0\linewidth}
\caption{\small Comparisons among different benchmarks. }
\begin{adjustbox}{scale=0.85,tabular= lccccccc,center}
  \begin{tabular}{cccccccccccc}
  \toprule  
Settings & CF  & Unlabeled   & Few-Shot  &  Unseen Tasks  \\ \midrule
CL \cite{EWC16} &\checkmark &\xmark &  \xmark & \xmark\\
SSL \cite{berthelot2019mixmatch} & \xmark  &  \checkmark &  \xmark & \xmark \\
CFSL \cite{antoniou2020defining} &\checkmark& \xmark & \checkmark & \xmark\\
SSCL \cite{wang2021ordisco} & \checkmark &\checkmark & \xmark  &  \xmark\\
SSFSL \cite{ren2018metalearning}& \xmark& \checkmark & \checkmark & \checkmark \\
 SETS (Ours) & \checkmark  &\checkmark  & \checkmark & \checkmark \\
    \bottomrule
  \end{tabular}
\label{tab:benchmark}
\end{adjustbox}
\end{wraptable}

We summarize and compare different benchmarks in Table \ref{tab:benchmark}, including continual learning (CL), semi-supervised learning (SSL), continual few-shot learning (CFSL),
semi-supervised continual learning (SSCL), semi-supervised few-shot learning (SSFSL), and SETS (Ours) in terms of whether they consider catastrophic forgetting (CF), use unlabeled data, few-shot learning, and generalize to unseen tasks. In summary, our setting is more comprehensive and practical and covers more learning aspects in real applications.

\section{Problem Setting}\label{sec:problemsetting}

\begin{figure}[t!] 
     \centering
     \begin{subfigure}[b]{0.48\textwidth}
         \centering
         \includegraphics[width=\textwidth]{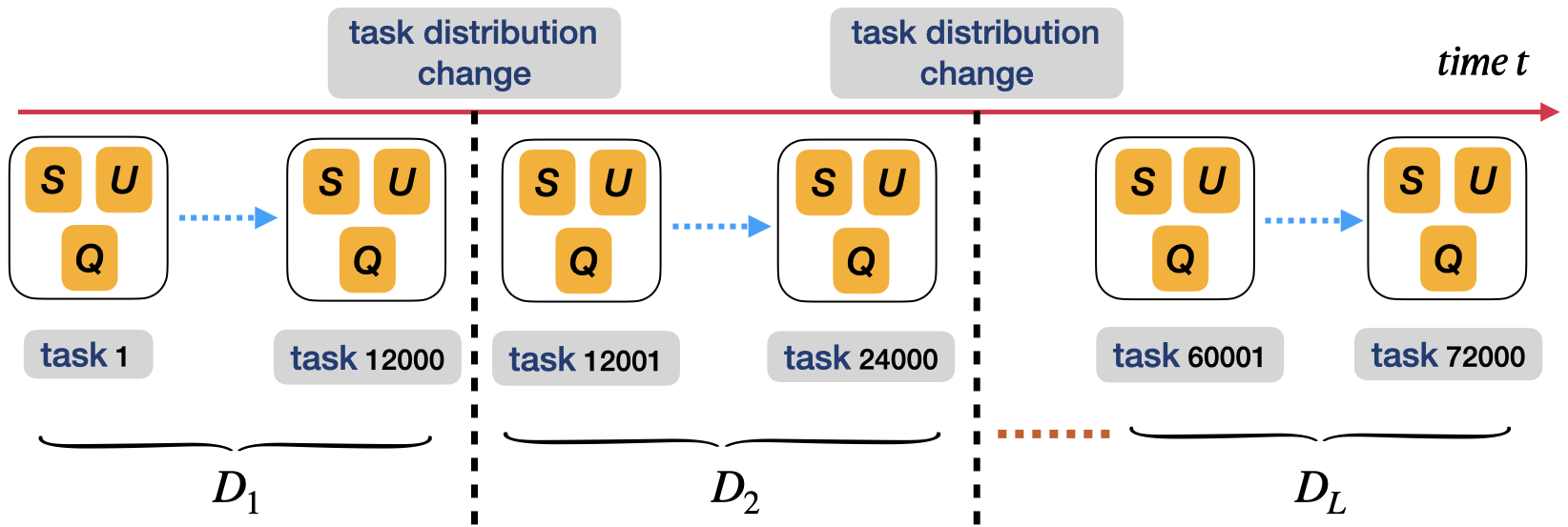}
     \end{subfigure}
     \hfill
     \begin{subfigure}[b]{0.48\textwidth}
         \centering
         \includegraphics[width=\textwidth]{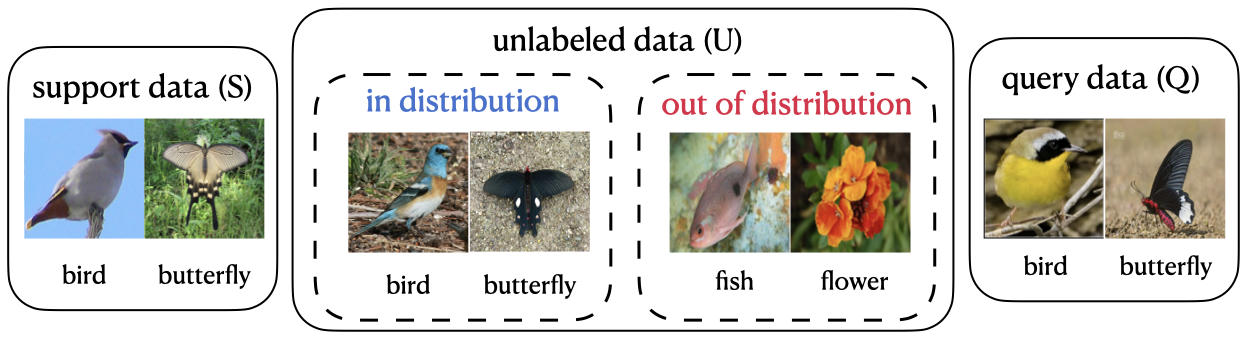}
     \end{subfigure}
     \caption{Illustration of SETS on a large-scale non-stationary semi-supervised task distributions $\mathcal{D}_1, \mathcal{D}_2, \cdots,  \mathcal{D}_L$ (top) with at least 72K tasks. Each task (bottom) contains labeled support data (S), unlabeled data (U) (in-distribution and out-of-distribution data) and query data (Q).}
    \label{fig:overview}
\end{figure}

\textbf{General setup.} SETS (Figure \ref{fig:overview} for illustration), online meta-learns on a sequence of task distributions, ${\mathcal{D}}_{1}, {\mathcal{D}}_{2}, \dots, {\mathcal{D}}_{L}$. For time $t = 1, \cdots, \tau_1$, we randomly sample mini-batch semi-supervised tasks $\mathcal{T}_t$ at each time $t$ from task distribution $P(\mathcal{D}_1)$; for time  $t = \tau_1+1, \cdots, \tau_2$, we randomly sample mini-batch semi-supervised tasks $\mathcal{T}_t$ at each time $t$ from task distribution $P(\mathcal{D}_2)$; for time $t = \tau_i+1, \cdots, \tau_{i+1}$, we randomly sample mini-batch semi-supervised  tasks $\mathcal{T}_t$ at each time $t$ from task distribution $P(\mathcal{D}_i)$, where $P(\mathcal{D}_i)$ is the task distribution (a collection of a large number of tasks) in $\mathcal{D}_i$. Thus, in SETS, each task from each domain sequentially arrives. Time $t = \tau_1, \tau_2, \cdots, \tau_i, \cdots$ is the time when task distribution shift happens, but we do not assume any prior knowledge about them. This is a more practical and general setup.  Each time interval $|\tau_i-\tau_{i-1}|$ is large enough to learn a large number of tasks from each task distribution adequately. 
 Each task $\mathcal{T}$ is divided into support, unlabeled and query data 
$\{\mathcal{S}, \mathcal{U}, \mathcal{Q} \}$. The support set $\mathcal{S}$ consists of a collection of labeled examples,  $\{(\xb^k, y^k)\}$, where $x^k$ is the data and $y^k$ is the label. The unlabeled set $\mathcal{U}$ contains only
inputs: $\mathcal{U} = \{\tilde{\xb}^1, \tilde{\xb}^2, \cdots, \}$. The goal is to predict labels for the examples in the task’s
query set $\mathcal{Q}$. Our proposed learning system maintains a memory buffer $\mathcal{M}$ to store a small number of training tasks from previous task distributions. We use reservoir sampling \cite{reservoirsampling} (RS) to maintain tasks in the memory. RS assigns equal probability for each incoming task without needing to know the total number of training tasks in advance. The total number of meta training tasks is much larger than the capacity of memory buffer, making it infeasible to store all the tasks. More details about memory buffer update and maintenance are provided in Appendix \ref{app:implementdetails}. 
We evaluate model performance on  the unseen tasks sampled from both current and all previous task distributions.

\textbf{Evolving semi-supervised few-shot episode distributions construction}. In most existing works, the meta-learning model is trained in a stationary setting. By contrast, in SETS, the model is meta trained on an evolving sequence of datasets. A large number of semi-supervised tasks sampled from each dataset  form task distribution $\mathcal{D}_i$. We thus have \textit{a sequence of semi-supervised task distributions} $\mathcal{D}_1, \mathcal{D}_2, \cdots, \mathcal{D}_L$. In episodic meta-learning, a task is usually denoted as an episode. When training on task distribution $\mathcal{D}_i$, to sample an $N$-way $K$-shot training episode, we first uniformly sample $N$ classes from the set of training classes $\mathcal{C}^{train}_i$. For the labeled support set $\mathcal{S}$ (training data), we then
sample $K$ images from the labeled split of each of these classes. For the unlabeled set $\mathcal{U}$, we sample $Z$
images from the unlabeled split of each of these classes as ID data. 
When including OOD in $\mathcal{U}$, we additionally sample $R$ images from external datasets as OOD. 
In practical scenarios, the OOD data could dominate the unlabeled portion $\mathcal{U}$. To be more realistic, we construct the unlabeled set with different proportions of OOD to simulate real-world scenarios. The query data (testing data)  $\mathcal{Q}$ of the episode is comprised of a fixed number of images from the labeled split of each of the $N$ chosen classes. The support set $\mathcal{S}$ and query set $\mathcal{Q}$ contain different inputs but with the same class set.

At the end of meta training on task distribution $\mathcal{D}_L$, the meta trained model is expected to generalize to the $unseen$ tasks of each dataset in the sequence. The evaluation on $unseen$ classes of previous datasets is to measure the forgetting. We adopt the $N$-way $K$-shot
classification on $unseen$ classes of each dataset. The semi-supervised testing episodes are constructed similarly as above. The accuracy is averaged on the query set $\mathcal{Q}$ of many
meta-testing episodes from all the trained datasets.

\textbf{Constructed dataset.}
To simulate realistic evolving semi-supervised task distributions, we construct a new large-scale dataset and collect 6 datasets, including \textbf{CIFARFS} \cite{bertinetto2019metalearning}, \textbf{AIRCRAFT} \cite{maji2013finegrained}, \textbf{CUB} \cite{WelinderEtal2010}, \textbf{Miniimagenet} \cite{matching16}, \textbf{Butterfly} \cite{chen2018finegrained} and \textbf{Plantae} \cite{vanhorn2018inaturalist}. The OOD data for each task are sampled from another three datasets, including \textbf{Omniglot} \cite{Omniglot2011}, \textbf{Vggflower} \cite{Nilsback08} and \textbf{Fish} \cite{zhuang2018wildfish}.  For each dataset, we randomly sample 12K tasks, we thus have 72K tasks in total. This task sequence is much larger than existing continual learning models considered. We perform standard 5-way-1-shot and 5-way-5-shot learning on this new dataset. All the datasets are publicly available with more details provided in Appendix \ref{appendix:dataset} due to the space limitation.

\section{Methodology}
In the following, we focus on the $N$-way $K$-shot setting of SETS as described above. We first describe the standard SSFSL task in Section \ref{sec:pre}, then present our method for handling unlabeled data, especially OOD data in Section \ref{sec:unlabel}. Next, we present our method for mitigating catastrophic forgetting in Section \ref{sec:forgetting}. At last, we summarize our proposed methodology in Section \ref{sec:total-loss}.

\subsection{Standard Semi-Supervised Few-Shot Learning}\label{sec:pre}

Prototypical network \cite{protonet17} has been extended to semi-supervised few shot learning by \cite{ren2018metalearning}. We denote the feature embedding function as $h_{\thetab}(\xb)$ with parameter $\thetab$. The class-wise prototype for support set $\mathcal{S}$ is calculated as : 
$\pb_c ={\sum_{\xb \in \mathcal{S}_c}  h_{\thetab}(\xb)}{\big/}{\sum_{\xb \in \mathcal{S}_c} 1 },$
where $\mathcal{S}_c$ is the set of data in $\mathcal{S}$ belonging to category $c$.  Each class prototype is refined by unlabeled data as: 
$$\pb_c^{\prime} = \frac{\sum_{\xb \in \mathcal{S}_c}  h_{\thetab}(\xb)+ \sum_{\tilde{\xb} \in \mathcal{U}}  h_{\thetab}(\tilde{\xb})  \mub_{c}(\tilde{\xb}) }{\sum_{\xb \in \mathcal{S}_c} 1 + \sum_{\tilde{\xb} \in \mathcal{U}} \mub_{c}(\tilde{\xb})}$$ 
by assigning each unlabeled data  $\tilde{\xb} \in \mathcal{U}$ with a soft class probability $ \mub_{c}(\tilde{\xb}) = \frac{exp(-||h_{\thetab}(\tilde{\xb})- \pb_c||_2^2)}{\sum_{c^{\prime}} exp(-||h_{\thetab}(\tilde{\xb})- \pb_c^{\prime}||_2^2)} $. To work with OOD data, they use an additional class
with zero mean in addition to the normal $c$ classes in the support set. The learning objective is to simply maximize the average log-probability of the correct class assignments over query examples $\mathcal{Q}$ and rigorously defined as following:
\begin{align}\label{eq:semimeta}
\small
     \mathcal{L}_{meta} 
     &= P(\mathcal{Q}|\mathcal{\thetab}, \mathcal{S}; \mathcal{U}) =  P(\mathcal{Q}|\mathcal{\thetab}, \{\pb_c^{\prime}\}) = {\sum}_{ (\xb_i, y_i) \in \mathcal{Q}} P(y_i|\xb_i, \mathcal{\thetab}, \{\pb_c^{\prime}\}).
\end{align}

\subsection{Mutual-Information For Unlabeled Data Handling}\label{sec:unlabel}

To be more realistic, the unlabeled data contains OOD data for each task, whose data categories are not present in the support set. In practice, the unlabeled OOD data could potentially negatively impact the performance. It has been shown that existing works \cite{ren2018metalearning,li2019learning,liu2019learning} are particularly sensitive to OOD data. How to properly use unlabeled data is challenging when OOD data dominates the unlabeled data, which is common in practical scenarios. Instead of sidestepping this problem by simply discarding the detected OOD data, we propose to exploit discriminative information implied by unlabeled OOD samples.
To explicitly address this problem, we propose to learn the representations that are robust to OOD data from an information-theoretical perspective.  We first give the detailed definitions for in-distribution (ID) and out-of-distribution (OOD) below.

\begin{definition}
Suppose a task $\mathcal{T}$ with data $\{\mathcal{S}, \mathcal{U}, \mathcal{Q} \}$, contains labeled categories $\mathcal{C}$ in $\mathcal{S}$. For unlabeled data $\mathcal{U}$, in-distribution unlabeled data is defined as the set of data $\xb$ whose categories $c \in \mathcal{C}$, i.e., $\mathcal{U}_{id} = \{\xb|\xb \in \mathcal{U}, c\in \mathcal{C}\}$; similarly, the out-of-distribution data is defined as the set of data $\xb$ whose categories $c \notin \mathcal{C}$, i.e., $\mathcal{U}_{ood}=\{\xb|\xb \in \mathcal{U},  c\notin \mathcal{C}\}$.
\end{definition}

\begin{wrapfigure}{r}{0.63\textwidth}
\begin{minipage}{1\linewidth}
 \begin{algorithm}[H]
  \small
	\caption{ OOD sample detection.}
	\label{alg:ooddetect}
	\begin{algorithmic}[1]
		\REQUIRE A task data  $\{\mathcal{S}, \mathcal{U},  \mathcal{Q} \}$; unlabeled query $\mathcal{Q}$ serves as ID-calibration set;
    	\STATE calculate prototypes $\pb_c$ for each class $c$  with support data $\mathcal{S}$.
    	\FOR{$\xb_i \in $ $\mathcal{Q}$ }
    	\STATE $\wb_i = h_{\thetab}(\xb_i)/||h_{\thetab}(\xb_i)||_2$ (normalization)
    	\STATE $d_i = d(\wb_i, \{\pb_c\}_{c=1}^N) \overset{def}{=} \min_{c} d(\wb_i, \pb_c)$
    	\ENDFOR
    	
    	\STATE thresh = $\mu(\{d_i\})+ \sigma(\{d_i\})$ [$\mu(\{d_i\})$ is the mean of $\{d_i\}$ and $\sigma(\{d_i\})$ is the standard deviation of $\{d_i\}$]
    	
    	\FOR{$\xb_i \in $ $\mathcal{U}$ }
    	\STATE $\wb_i = h_{\thetab}(\xb_i)/||h_{\thetab}(\xb_i)||_2$ (normalization)
    	\STATE $d_i = d(\wb_i, \{\pb_c\}_{c=1}^N) \overset{def}{=} \min_{c} d(\wb_i, \pb_c)$
    	\IF{$d_i$>thresh}
    	\STATE $\mathcal{U}_{ood} = \mathcal{U}_{ood}\bigcup \xb_i$
    	\ELSE
    	\STATE $\mathcal{U}_{id} = \mathcal{U}_{id}\bigcup \xb_i$
    	\ENDIF
    	\ENDFOR
    	\STATE return ID data $\mathcal{U}_{id}$ and OOD data $\mathcal{U}_{ood}$ 
	\end{algorithmic}
\end{algorithm}
\end{minipage}
\end{wrapfigure} 
\textbf{OOD sample detection.}\ We propose to automatically divide the unlabeled data for each task into ID and OOD. The existing confidence-based approach is highly unreliable for detecting OOD data in the few-shot learning setting since the network can produce incorrect high-confidence predictions for OOD samples \cite{nguyen2015deep}. We instead propose a metric-based OOD sample detection method in feature embedding space. First, we calculate the class-wise prototypes $\pb_c$ for each class $c$. We define the distance of data $\xb$ to the prototypes set as: $d(h_{\thetab}(\xb), \{\pb_c\}_{c=1}^N) \overset{def}{=} \min_{c} d(h_{\thetab}(\xb), \pb_c)$, i.e., the smallest distance to all the prototypes in the embedding space; where we use Euclidean distance as the distance metric $d$ for simplicity. Intuitively, the data closer to the class prototypes are more likely to be ID data,  the data farther to the class prototypes are more likely to be OOD.  We adopt a method similar to the state-of-the-art few-shot OOD detection algorithm, named cluster-conditioned detection \cite{sehwag2021ssd}. We use the unlabeled query data $\mathcal{Q}$ for each task as an ID calibration set to determine the threshold between OOD and ID data. The OOD sample detection is 
presented in Algorithm \ref {alg:ooddetect}. The threshold is based on the Gaussian assumption of 
$d(h_{\thetab}(\xb), \{\pb_c\}_{c=1}^N)$ (3-Sigma rule).

We adopt mutual information (MI) to separately handle ID data and OOD data, where MI describes the mutual dependence between the two variables and is defined as follows. 
\begin{definition}
  The mutual information between variables $\ab$ and $\bb$ is defined as 
\begin{equation}\label{eq:MI}
\small
    \mathcal{I}(\ab, \bb) = \mathbb{E}_{p(\ab, \bb)} \left[\log \frac{p(\ab, \bb)}{p(\ab) p(\bb)}\right],
\end{equation}
where $p(\ab)$ and  $p(\bb)$ are marginal distribution of $\ab$ and $\bb$, $p(\ab, \bb)$ is the joint distribution.
\end{definition}

In information theory, $\mathcal{I}(\ab, \bb)$ is a measure of the mutual dependence between the two variables $\ab$ and $\bb$. More specifically, it quantifies the "amount of information" about the variable $\ab$ (or $\bb$ ) captured via observing the other variable $\bb$ (or $\ab$ ). The higher dependency between two variables, the higher MI is. Intuitively, the higher dependency between embedded OOD data and class prototypes, the higher the negative impact on the model performance. On the contrary, more dependency between embedded ID data and class prototypes could improve the performance. To enhance the effect of ID data on the class prototypes, we maximize the MI between class prototypes and ID data. To reduce the negative effect of OOD data on the class prototypes, we further consider minimizing the MI between class prototypes and OOD data:
\begin{equation}\label{eq:meta+mi}
\small
    \mathcal{L}_{total} =  \mathcal{L}_{meta} +  \lambda\left(\mathcal{I} (\eb_{ood}, \pb_{c^{\prime}}) - \mathcal{I} (\eb_{id}, \pb_{c^{\prime}})\right),
\end{equation}
where $\mathcal{L}_{meta}$ is defined in Eq. \eqref{eq:semimeta};  $\eb_{ood} = h_{\thetab}(\xb), \xb\in \mathcal{U}_{ood}$ and $\eb_{id} = h_{\thetab}(\xb), \xb\in \mathcal{U}_{id}$. $c^{\prime} =  \argmin_{c}  ||\eb - \pb_c||$; $\mathcal{I} (\eb_{ood}, \pb_{c^{\prime}})$ is the mutual information between embedded OOD data and class prototypes; $\mathcal{I} (\eb_{id}, \pb_{c^{\prime}})$ is the mutual information between embedded ID data and class prototypes; $\lambda$ is a hyperparameter to control the relative importance of different terms. Since this objective is to minimize $-\mathcal{I} (\eb_{id}, \pb_{c^{\prime}})$, it is equivalent to maximize   $\mathcal{I} (\eb_{id}, \pb_{c^{\prime}})$. However, the  computation of MI
values is intractable \cite{poole2019variational,pmlrv119cheng20b}, since the form of joint probability distribution $p(\ab, \bb)$ in Eq. \eqref{eq:MI}
is not available in our case and only samples from the
joint distribution are accessible. To solve this problem, we derive the variational lower bound of MI for MI maximization  and variational upper bound of MI for MI minimization respectively. Due to the limited space, all proofs for Lemmas and Theorems are provided in Appendix \ref{app:theory}.

\begin{theorem}\label{theorem:upper}
For a task $\mathcal{T} = \{\mathcal{S}, \mathcal{U}, \mathcal{Q}\} $, suppose the unlabeled OOD data embedding $\eb_{ood} = h_{\thetab}(\xb), \xb\in \mathcal{U}_{ood}$. The nearest prototype corresponds to $\eb_{ood}$ is $\pb_{c^{\prime}}$, where $c^{\prime} =  \argmin_{c}  ||\eb_{ood} - \pb_c||$. Given a collection of samples $\{(\eb_{ood}^{i}, \pb_{c^{\prime}}^{i})\}_{i=1}^{i=L} \sim P(\eb_{ood}, \pb_{c^{\prime}})$, the  variational upper bound of mutual information $ \mathcal{I} (\eb_{ood}, \pb_{c^{\prime}})$ is: 
 \begin{align}\label{eq:ood}
 \small
      \mathcal{I} (\eb_{ood}, \pb_{c^{\prime}}) 
      \leq \sum_{i=1}^{i=L} \log P(\pb_{c^{\prime}}^i|\eb_{ood}^i) -  \sum_{i=1}^{i=L}\sum_{j=1}^{j=L} \log P(\pb_{c^{\prime}}^i|\eb_{ood}^j) 
      = \mathcal{I}_{ood}.
 \end{align}
 The bound is tight (equality holds) when $\pb_{c^{\prime}}$ and $\eb_{ood}$ are independent.
\end{theorem}

This upper bound requires the unknown conditional probability distribution $P(\pb_{c^{\prime}}|\eb_{ood})$ and thus it is intractable. In this work, we approximate it with variational distribution   $Z_{\phi}(\pb_{c^{\prime}}|\eb_{ood})$ parameterized by a neural network.

\begin{lemma}\label{eq:lemma}
 For a task $\mathcal{T} = \{\mathcal{S}, \mathcal{U}, \mathcal{Q}\} $, suppose the  unlabeled ID data embedding $\eb_{id} = h_{\thetab}(\xb), \xb\in \mathcal{U}_{id}$. The nearest prototype corresponds to $\eb_{id}$ is $\pb_{c^{\prime}}$, where $c^{\prime} =  \argmin_{c}  ||\eb_{id} - \pb_c||$. Given a collection of samples $\{(\eb_{id}^{i}, \pb_{c^{\prime}}^{i})\}_{i=1}^{i=L} \sim P(\eb_{id}, \pb_{c^{\prime}})$, the variational lower bound of $\mathcal{I} (\eb_{id}, \pb_{c^{\prime}})$ is:
 \begin{equation}
 \small
       \mathcal{I} (\eb_{id}, \pb_{c^{\prime}}) \geqslant \sum_{i=1}^{i=L} f(\eb_{id}^i, \pb_{c^{\prime}}^i) - \sum_{j=1}^{j=L}\log \sum_{i=1}^{i=L} e^{f(\eb_{id}^{i}, \pb_{c^{\prime}}^j)}.
 \end{equation}
 This bound is tight if $f(\eb_{id}, \pb_{c^{\prime}}) = \log P(\pb_{c^{\prime}}|\eb_{id}) + c(\pb_{c^{\prime}})$.
\end{lemma}

Based on Lemma \ref{eq:lemma}, the variational lower bound is derived as in Theorem \ref{theorem:lower}.

\begin{theorem}\label{theorem:lower}
 For a task $\mathcal{T} = \{\mathcal{S}, \mathcal{U}, \mathcal{Q}\} $, suppose the unlabeled ID data embedding $\eb_{id} = h_{\thetab}(\xb), \xb\in \mathcal{U}_{id}$. The nearest prototype corresponds to $\eb_{id}$ is $\pb_{c^{\prime}}$, where $c^{\prime} =  \argmin_{c}  ||\eb_{id} - \pb_c||$. Given a collection of samples $\{(\eb_{id}^{i}, \pb_{c^{\prime}}^{i})\}_{i=1}^{i=L} \sim P(\eb_{id}, \pb_{c^{\prime}})$, the variational lower bound of $\mathcal{I} (\eb_{id}, \pb_{c^{\prime}})$ is:
 \begin{align}\label{eq:id}
 \small
    \!\!\!\! \mathcal{I} (\eb_{id}, \pb_{c^{\prime}}) \!\geqslant \! 
     \mathcal{I}_{id} \!:=\!\sum_{i=1}^{i=L} f(\eb_{id}^i, \pb_{c^{\prime}}^i)   \! -\! \sum_{j=1}^{j=L} \left[\frac{\sum_{i=1}^{i=L}  e^{f(\eb_{id}^i, \pb_{c^{\prime}}^j)} }{a(\pb_{c^{\prime}}^j)} \!+\! \log(a(\pb_{c^{\prime}}^j))-1\right].   
 \end{align}
 The bound is tight (equality holds) when $f(\eb_{id}, \pb_{c^{\prime}}) = \log P(\pb_{c^{\prime}}|\eb_{id}) + c(\pb_{c^{\prime}})$ and $a(\pb_{c^{\prime}}) = \mathbb{E}_{p(\eb_{id})} e^{f(\eb_{id}, \pb_{c^{\prime}})}$. $a(\pb_{c^{\prime}})$ is any function that  $a(\pb_{c^{\prime}})>0$.
\end{theorem}
 
Hence, by combining Theorems \ref{theorem:upper} and \ref{theorem:lower}, we obtain an surrogate for the loss function Eq.~\eqref{eq:meta+mi} as:
\begin{align}\label{Eq:tractable-meta+mi}
    \mathcal{L}_{total} 
    &=  \mathcal{L}_{meta} +  \lambda(\mathcal{I} (\eb_{ood}, \pb_{c^{\prime}}) - \mathcal{I} (\eb_{id}, \pb_{c^{\prime}})) \leq \mathcal{L}_{meta} +  \lambda(\mathcal{I}_{ood} - \mathcal{I}_{id}).
\end{align}

\subsection{Mitigate CF by Optimal Transport}\label{sec:forgetting}

A straightforward idea of mitigating catastrophic forgetting on previous task distributions is to replay the memory tasks when training on current task $\mathcal{T}_t$, i.e., we randomly sample a small batch tasks $\mathcal{B}$ from the memory buffer $\mathcal{M}$. Thus the loss function becomes $\mathcal{H}(\thetab) = \mathcal{L}(\mathcal{T}_t) + \sum_{\mathcal{T}_i\in \mathcal{B}} \mathcal{L}(\mathcal{T}_i)$. However, simple experience replay only forces the model to remember a limited number of labeled data, i.e., a much larger number of previous unlabeled data is totally neglected, which causes the model to sub-optimally remember the previous knowledge. To further avoid forgetting, we directly regularize the network outputs on both labeled and unlabeled data since what ultimately matters is the network output, in addition to remembering output labels on memory tasks.
To achieve this goal, we propose a novel constrained loss via optimal transport to reduce the \textit{output (feature) distribution shift} on both labeled and unlabeled data. 

{\bf Optimal Transport.}\ 
A discrete distribution $\bm{\mu} = \sum_{i=1}^n u_i\delta_{\eb_i}(\cdot)$, where $u_i \geq 0$ and $\sum_iu_i = 1$, $\delta_{\eb}(\cdot)$ denotes a spike distribution located at $\eb$. Given two discrete distributions, $\bm{\mu} = \sum_{i=1} ^ {n} u_{i}  \delta_{\eb_i}$ and $\bm{\nu} = \sum_{j=1} ^ {m} v_{j}  \delta_{\bm{g}_j}$, respectively, the OT distance \cite{NEURIPS2020_ROT} between $\bm{\mu}$ and $\bm{\nu}$ is defined as the optimal value of the following problem: 
\begin{equation}\label{eq:OTloss}
\small
 \mathcal{L}_{\text{OT}} =   { \underset {\bm{W}  \in \Pi(\bm{\mu}, \bm{\nu})} {\min}}  \sum_{i=1} ^{n} \sum_{j=1} ^{m} \bm{W}_{ij} \cdot d(\bm{e}_i,\bm{g}_j)~,
\end{equation}
where $\bm{W}_{ij}$ is defined as the joint probability mass function on $(\bm{e}_i,\bm{g}_j)$ and $d(\mathbf{e},\mathbf{g})$ is the cost of moving $\mathbf{e}$ to $\mathbf{g}$ (matching $\mathbf{e}$ and $\mathbf{g}$). $\Pi(\bm{\mu}, \bm{\nu})$ is the set of joint probability distributions with two marginal distributions equal to $\bm{\mu}$ and $\bm{\nu}$, respectively.  
The exact calculation of the OT loss is generally difficult \cite{Genevay2018}. Therefore, for simplicity, we use CVXPY \cite{cvxpylayers2019} for efficient approximation.

\begin{wrapfigure}{r}{0.65\textwidth}
\begin{minipage}{1\linewidth}
 \begin{algorithm}[H]
  \small
	\caption{\ ORDER Algorithm}
	\label{alg:selectunlabel}
	\begin{algorithmic}[1]
		\REQUIRE evolving episodes ${\mathcal{T}}_{1}, {\mathcal{T}}_{2},\dots, {\mathcal{T}}_{N}$, with $\{\tau_1, < \cdots, \tau_i, \cdots, < \tau_{L-1} \}$ the time steps when task distribution shift;  memory buffer $\mathcal{M} \!=\! \{\}$; model parameters $\thetab$; memory task features $\mathcal{E} \!=\! \{\}$. \!\!\!
		\FOR{$t = 1$ to $N$}
    \STATE sample tasks $\mathcal{B}$ from  $\mathcal{M}$ and $\mathcal{B} = \mathcal{B} \bigcup \mathcal{T}_t$.
    	\FOR{$\mathcal{T} \in $ $\mathcal{B}$ }
    	\STATE $\mathcal{T} = \{\mathcal{S}, \mathcal{U}, \mathcal{Q} \}$, 
    	    OOD detection to divide the unlabeled data $\mathcal{U}$ into $\mathcal{U}_{ood}$ and $\mathcal{U}_{id}$ 
    	    \STATE update parameters $\thetab$ by minimizing $\mathcal{H}(\thetab)$
    	\ENDFOR
    	\IF{reservoir sampling:}
    	 \STATE  $\mathcal{M} = \mathcal{M} \bigcup \mathcal{T}_t$ // store raw task data
    	 \STATE  $\mathcal{E} =  \mathcal{E} \bigcup \mathcal{E}(\mathcal{T}_t)$ // store task data features
    	 \ENDIF
		\ENDFOR
	\end{algorithmic}
\end{algorithm}
\end{minipage}
\end{wrapfigure}

When storing a semi-supervised episode into the memory, we store the feature embedding for each data point in addition to the raw data. Memory features are the collection of $\rm \{h_{\thetab}(\xb), \xb \in \bigcup_{\{\mathcal{S},  \mathcal{U}_{id} \} \in \mathcal{M}} \{\mathcal{S}, \mathcal{U}_{id}\} \}$, the concatenation of labeled and unlabeled ID data embedding. Suppose the previously stored data features are $\{ \bm{g}_i \}_{i=1}^G$ with distribution $\bm{\nu}$, where $G$ is the number of data examples stored in memory buffer. The data features generated with model parameters $\thetab_t$ of current iteration are  $\{ \bm{e}_i \}_{i=1}^G$ with distribution $\bm{\mu}$. Then, the loss function becomes 
minimizing on the optimal transport between $\{ \bm{e}_i \}_{i=1}^G$ and $\{ \bm{g}_i \}_{i=1}^G$.

\subsection{Overall Learning Objective}\label{sec:total-loss}

By unifying the above-derived novel mutual information driven unlabeled data handling technique and optimal transport driven forgetting mitigation, we propose an OOD robust and knowledge preserved semi-supervised meta-learning algorithm (ORDER) to tackle the challenges in the SETS setting. 
Combining the OT loss from Eq.~\eqref{eq:OTloss} with Eq.~\eqref{Eq:tractable-meta+mi} and assigning $\lambda$ and $\beta$ as two hyperparameters for controlling the relative importance of the corresponding terms, 
we obtain the following final loss function:
\begin{align} \label{eq:totalloss}
\small
&\mathcal{H}(\thetab) 
   = \mathcal{L}_{meta}(\mathcal{T}_t) + \lambda (\mathcal{I}_{ood} - \mathcal{I}_{id})+ \sum \limits_{\mathcal{T}_i\in \mathcal{B}} \mathcal{L}_{total}(\mathcal{T}_i) + \beta \mathcal{L}_{\text{OT}}.
\end{align}
The benefits of the second term are two-fold: (i) promote the effect of ID data on class prototypes by maximizing the MI between class prototypes and ID data; (ii) reduce the negativity of OOD data on class prototypes by minimizing the MI between class prototypes and OOD data. The last term is to force the model to remember in feature space. 
The detailed procedure of ORDER is shown in Algorithm \ref{alg:selectunlabel}.

\section{Experiments}

\begin{wraptable}{r}{0.6\textwidth}
\small
\captionsetup{width=0.9\linewidth}
\caption{5-way, 1-shot and 5-shot  accuracy compared to meta-learning baselines.}
\begin{adjustbox}{scale=0.85,tabular= lccccccc,center}
\begin{tabular}{lrrrrrrrrr}
\toprule
Algorithm & 1-Shot &  5-Shot &\\
\midrule  
ProtoNet  &$31.96\pm 0.93$ &  $42.62\pm 0.73$ & \\
ANIL  &$30.58\pm 0.81$ &  $42.24\pm 0.86$ & \\
\midrule
MSKM  &$33.79\pm 1.05$  &  $45.41\pm 0.58$  \\
LST & $34.81\pm 0.81$  &  $46.09\pm 0.69$ &  \\
TPN  &$34.52\pm 0.83$  &  $46.23\pm 0.60$  \\
ORDER (Ours with only MI) &$36.93\pm 0.90$ & $48.85\pm 0.53$  \\
ORDER (Ours) &$\bm{41.25\pm 0.64}$ & $\bm{53.96\pm 0.71}$   \\
\bottomrule
\end{tabular}
\label{tab:supervisedbaseline1}
\end{adjustbox}
\end{wraptable}

To show  the benefit of using unlabeled data, we first compare semi-supervised meta-learning using unlabeled data with meta-learning without unlabeled data in Section \ref{unlabel}. To evaluate the effectiveness of the proposed method, ORDER,   for learning useful knowledge in SETS, we compare to SOTA meta-learning models in Section \ref{ml-compare}.  Next, we compare ORDER to SOTA continual learning baselines to show the effectiveness of ORDER for mitigating forgetting in SETS in Section \ref{cl-compare}.  
For the evaluation, we calculate the average testing accuracy on previously unseen 600 testing tasks sampled from the unseen categories of each dataset in the dataset sequence to evaluate the effectiveness of the tested methods. Moreover, we conduct extensive ablation studies in Section \ref{ablation}. 
Due to the space limitation, we put detailed implementation details in Appendix \ref{app:implementdetails}.

 \subsection{Benefit of Using Unlabeled Data in SETS}\label{unlabel}

To verify the benefit of using unlabeled data in SETS, we compare with various meta-learning methods including supervised and semi-supervised ones. All of these methods are adapted to SETS setting directly. Supervised meta-learning methods \textit{without unlabeled data}, including 
(1) gradient-based meta-learning \textbf{ANIL} \cite{raghu2020rapid}, which is a simplified model of MAML \cite{finn17a}; (2) metric-based meta-learning \textbf{Prototypical Network} (\textbf{ProtoNet}) \cite{protonet17}. 
 In addition, we compare to representative  SSFSL models \textit{using unlabeled data}: (3) Masked Soft k-Means (\textbf{MSKM}) \cite{ren2018metalearning}; (4) Transductive Propagation Network (\textbf{TPN}) \cite{liu2019learning}; (5) Learning to Self-Train (\textbf{LST}) \cite{li2019learning}. LST and MSKM can be viewed as extension of MAML(ANIL) and ProtoNet to SETS by using additional unlabeled data respectively. Table \ref{tab:supervisedbaseline1} shows that with additional unlabeled data, semi-supervised meta-learning methods substantially outperform corresponding meta-learning methods without using unlabeled data, demonstrating that unlabeled data could further improve performance in SETS. We also include a reduced version of our method, which uses Eq.~\eqref{Eq:tractable-meta+mi} as the objective function, and keep other components the same as baselines without considering mitigating CF. We can see that with only the guidance of MI, our method can outperform baselines by more than 2.6\% for 5-shot learning and 2.1\% for 1-shot learning, which verifies the effectiveness of the proposed MI for handling unlabeled data.

\subsection{Comparison to Meta-Learning}\label{ml-compare}

\begin{wraptable}{r}{0.58\textwidth}
\small
\centering
\captionsetup{width=0.9\linewidth}
\caption{5-way, 1-shot and 5-shot classification accuracy compared to continual learning baselines.}
\begin{adjustbox}{scale=0.85,tabular= lccccccc,center}
\begin{tabular}{lrrrrrrrrr}
\toprule
Algorithm & 1-Shot &  5-Shot &\\
\midrule 
Semi-Seq  &$33.79\pm 1.05$  &  $45.41\pm 0.58$  \\
Semi-ER  &$37.62\pm 0.97$  &  $50.35\pm 0.76$  \\
Semi-AGEM  &$36.97\pm 1.16$  &  $50.29\pm 0.69$  \\
Semi-MER  &$37.90\pm 0.83$  &  $50.46\pm 0.74$  \\
Semi-GPM & $36.53\pm 0.81$  &  $49.78\pm 0.65$ &  \\
Semi-DEGCL & $36.78\pm 0.89$  &  $50.07\pm 0.61$ &  \\
\midrule
ORDER (Ours OT only) & $39.21\pm 0.61$ & $51.72\pm 0.61$ \\
ORDER (Ours)  &$\bm{41.25\pm 0.64}$ & $\bm{53.96\pm 0.71}$   \\
\midrule
Joint-training & $49.91\pm 0.79$ &  $61.78\pm 0.75$\\
\bottomrule
\end{tabular}
\label{tab:protonetbaseline1}
\end{adjustbox}
\end{wraptable}

To compare to SOTA meta-learning methods, similar to the baselines in Section \ref{unlabel}, Table \ref{tab:supervisedbaseline1} also shows the advantage of ORDER. We observe that our method significantly outperforms baselines for 5-shot learning by $7.7\%$ and for 1-shot learning by  $6.4\%$.  This improvement shows the effectiveness of our method in this challenging setting. LST \cite{li2019learning} performs relatively worse in SETS, probably due to the challenging OOD data and noise of pseudo-labels.

\subsection{Comparison to Continual Learning}\label{cl-compare}
To evaluate the effectiveness of ORDER for mitigating forgetting in SETS, we compare to SOTA CL baselines, including Experience replay (ER) \cite{ERRing19},  A-GEM\cite{AGEM19}, MER \cite{riemer2018learning}, GPM \cite{saha2021gradient} and DEGCL \cite{buzzega2020dark}. Note that all of these methods are originally designed for mitigating forgetting for \textit{a small number of tasks}, we adapt these methods to SETS by combing them with meta-learning methods for mitigating forgetting. Here, we combine CL methods with MSKM for illustration since it is more efficient and performs comparably in SETS than the other two meta-learning methods. The combination baselines are named as Semi-, etc. We also compare to (1) \textbf{sequential training} (\textbf{Seq}), which trains the latent task distributions sequentially without any external mechanism and helps us understand the model forgetting behavior;  (2) \textbf{joint offline training}, which learns all the task distributions jointly in a stationary setting and provides the performance upper bound. 

The memory buffer is updated by reservoir sampling mentioned in Section \ref{sec:problemsetting}. More details about how to update the memory buffer is provided in Appendix \ref{app:implementdetails}. All the baselines and our method maintain 200 tasks in memory and the number of OOD data $R = 50$ for each task. Table \ref{tab:protonetbaseline1} shows the results. We also include a version of our method with memory replay and only OT regularization, without MI regularization, and other components are kept as the same as baselines. Our method can outperform baselines by $1.3\%$ for 1-shot learning and $1.2\%$ for 5-shot learning.
Our full method significantly outperforms baselines for 5-shot learning by $3.1\%$ and for 1-shot learning by  $3.3\%$.  This improvement shows the effectiveness of ORDER for mitigating forgetting in SETS.

\subsection{Ablation Study}\label{ablation}

\textbf{Robustness of OOD.} As for the OOD data of each episode, we randomly sample OOD from the mixture of the OOD dataset as described in Section \ref{sec:problemsetting}. Furthermore, it is more natural for the unlabeled set to include more OOD samples than ID samples. To investigate the sensitivity of the baselines and our method to OOD data, we fix the number of ID data for each task and gradually increase the amount of OOD data in the unlabeled set. We set  $R$ (number of OOD data) $= {50, 100, 150, 200, 250}$, respectively. The results are shown in Figure \ref{fig:ood} in Appendix \ref{sec:order}. Interestingly, we can observe that our method is much more robust to OOD data than baselines, even when the number of OOD data is large.

To study the individual effect of mutual information regularization and optimal transport, we conduct ablation studies to justify (i) whether adding mutual information regularization helps to be robust to OOD data;
 (ii) whether adding OT helps mitigate catastrophic forgetting. 
The results of both ablation studies are shown in Table \ref{tab:ablation} in Appendix \ref{sec:order}.  The \checkmark means the corresponding column component is used and \xmark means do not use the corresponding column component. Specifically, \quotes{MI} means using mutual information regularization, \quotes{OT} means using optimal transport constraint loss, \quotes{$\mathcal{U}_{id}$} means using additional ID unlabeled data features in addition to labeled data for OT loss. We observe that the improvement of robustness to OOD becomes more significant, especially with a larger amount of OOD data. 
When the number of OOD is 250 for each task, the improvement is nearly $4\%$. This demonstrates the effectiveness of mutual information regularization.  The optimal transport (OT) component improves model performance by more than $2\%$ with 250 OOD data, demonstrating the effectiveness of remembering in feature space with both labeled and unlabeled data. Additionally, we compare OT regularization with only labeled data features to baseline.  We find the performance is moderately improved in this case.  This shows the benefits of using unlabeled data for remembering.

To investigate whether using OOD data is helpful, we performed comparisons to cases where we only use ID-data for training and prediction with OOD data filtered out. 
The results are shown in Table \ref{tab:ablation2} in Appendix \ref{sec:order}. The results show that filtering out OOD data is already helpful, but with additional OOD data and MI regularization, the performance is improved further. 
Results on other ablation settings are also available in Table \ref{tab:ablation2}.

\textbf{Sensitivity to dataset ordering.}
To investigate the sensitivity of baselines and our method to dataset order, we also performed comparisons on two other dataset sequences, including (i) Butterfly, CUB, CIFARFS, Plantae, MiniImagenet, Aircraft and (ii) CUB, CIFARFS, Plantae, MiniImagenet, Butterfly, Aircraft. The results are shown in Appendix \ref{sec:order} (see details in Table \ref{tab:metabaseline2} and Table \ref{tab:continual2} for order (i), Table \ref{tab:metabaseline3} and Table \ref{tab:continual3} for order (ii)). In all cases, our method substantially outperforms the baselines and demonstrates its superiority.

\textbf{Sensitivity of hyperparameter.}
The sensitivity analysis of hyperparameters $\lambda$ and $\beta$ of ORDER are provided in Appendix \ref{sec:order} (see details in Table \ref{tab:hyper}). $\beta$ controls the magnitude of optimal transport regularization. Results indicate the model performance is positively correlated with $\beta$ when it increases until optimal trade-off is reached after which performance deteriorates with over regularization when $\beta$ reaches 0.01. A similar trend is observed in mutual information regularization $\lambda$.

\textbf{More Results.}
To investigate the effect of memory buffer size, we provide more experimental results with a memory buffer size of 50, 100, 200 tasks in Appendix \ref{sec:order} (see details in Table \ref{tab:memorysize}). With increasing memory buffer size, the performance improves further.  The current size of the memory buffer is negligible compared to the total amount of 72K tasks.

\section{Conclusion}

In this paper, we step towards a challenging and realistic problem scenario named SETS.  The proposed method ORDER is designed to leverage unlabeled data, especially OOD data, and to alleviate the forgetting of previously learned knowledge. Our method first detects unlabeled OOD data from ID data. Then with the guidance of mutual information regularization, it further improves the accuracy. The method is also shown to be less sensitive to OOD. The OT constraint is adopted to mitigate the forgetting in feature space. We have validated our methods on the constructed dataset. For the future work direction, we are working on adapting the method to the unsupervised scenario that no labeled data in new coming tasks, which requires the model with stronger robustness and capability of learning feature representations.

\textbf{Acknowledgement}   We thank all the anonymous reviewers for their thoughtful and insightful comments. This research was supported in part by NSF through grant IIS-1910492.

\clearpage
%
%
\bibliographystyle{splncs04}
\bibliography{egbib}

\begin{thebibliography}{10}
\providecommand{\url}[1]{\texttt{#1}}
\providecommand{\urlprefix}{URL }
\providecommand{\doi}[1]{https://doi.org/#1}

\bibitem{cvxpylayers2019}
Agrawal, A., Amos, B., Barratt, S., Boyd, S., Diamond, S., Kolter, Z.:
  Differentiable convex optimization layers. In: Advances in Neural Information
  Processing Systems (2019)

\bibitem{aljundi2018memory}
Aljundi, R., Babiloni, F., Elhoseiny, M., Rohrbach, M., Tuytelaars, T.: Memory
  aware synapses: Learning what (not) to forget. The 2018 European Conference
  on Computer Vision (ECCV)  (2018)

\bibitem{aljundi2019taskfree}
Aljundi, R., Kelchtermans, K., Tuytelaars, T.: Task-free continual learning.
  Proceedings of the IEEE Conference on Computer Vision and Pattern Recognition
  (CVPR)  (2019)

\bibitem{learn2learn2016}
Andrychowicz, M., Denil, M., Gomez, S., Hoffman, M.W., Pfau, D., Schaul, T.,
  Shillingford, B., de~Freitas, N.: Learning to learn by gradient descent by
  gradient descent. Advances in Neural Information Processing Systems  (2016)

\bibitem{antoniou2018train}
Antoniou, A., Edwards, H., Storkey, A.: How to train your maml. International
  Conference on Learning Representations  (2019)

\bibitem{antoniou2020defining}
Antoniou, A., Patacchiola, M., Ochal, M., Storkey, A.: Defining benchmarks for
  continual few-shot learning. https://arxiv.org/abs/2004.11967  (2020)

\bibitem{bae2020selfdriving}
Bae, I., Moon, J., Jhung, J., Suk, H., Kim, T., Park, H., Cha, J., Kim, J.,
  Kim, D., Kim, S.: Self-driving like a human driver instead of a robocar:
  Personalized comfortable driving experience for autonomous vehicles. NeurIPS
  Workshop  (2019)

\bibitem{NEURIPS2020_ROT}
Balaji, Y., Chellappa, R., Feizi, S.: Robust optimal transport with
  applications in generative modeling and domain adaptation. In: Advances in
  Neural Information Processing Systems. pp. 12934--12944 (2020)

\bibitem{VIbound}
Barber, D., Agakov, F.: The im algorithm: A variational approach to information
  maximization  (2003)

\bibitem{IL2MCV}
{Belouadah}, E., {Popescu}, A.: Il2m: Class incremental learning with dual
  memory. In: 2019 IEEE/CVF International Conference on Computer Vision (ICCV).
  pp. 583--592 (2019)

\bibitem{berthelot2019mixmatch}
Berthelot, D., Carlini, N., Goodfellow, I., Papernot, N., Oliver, A., Raffel,
  C.: Mixmatch: A holistic approach to semi-supervised learning  (2019)

\bibitem{bertinetto2019metalearning}
Bertinetto, L., Henriques, J.F., Torr, P.H.S., Vedaldi, A.: Meta-learning with
  differentiable closed-form solvers. International Conference on Learning
  Representations  (2019)

\bibitem{buzzega2020dark}
Buzzega, P., Boschini, M., Porrello, A., Abati, D., Calderara, S.: Dark
  experience for general continual learning: a strong, simple baseline. 34th
  Conference on Neural Information Processing Systems  (2020)

\bibitem{AGEM19}
Chaudhry, A., Ranzato, M., Rohrbach, M., Elhoseiny, M.: Efficient lifelong
  learning with a-gem. Proceedings of the International Conference on Learning
  Representations  (2019)

\bibitem{ERRing19}
Chaudhry, A., Rohrbach, M., Elhoseiny, M., Ajanthan, T., Dokania, P.K., Torr,
  P.H.S., Ranzato, M.: Continual learning with tiny episodic memories.
  https://arxiv.org/abs/1902.10486  (2019)

\bibitem{chen2018finegrained}
Chen, T., Wu, W., Gao, Y., Dong, L., Luo, X., Lin, L.: Fine-grained
  representation learning and recognition by exploiting hierarchical semantic
  embedding. ACM International Conference on Multimedia  (2018)

\bibitem{chen2019closerfewshot}
Chen, W.Y., Liu, Y.C., Kira, Z., Wang, Y.C., Huang, J.B.: A closer look at
  few-shot classification. In: International Conference on Learning
  Representations (2019)

\bibitem{pmlrv119cheng20b}
Cheng, P., Hao, W., Dai, S., Liu, J., Gan, Z., Carin, L.: Club: A contrastive
  log-ratio upper bound of mutual information. Proceedings of the 37th
  International Conference on Machine Learning  (2020)

\bibitem{CL2020}
Diethe, T.: Practical considerations for continual learning ({Amazon})  (2020)

\bibitem{ebrahimi2020adversarial}
Ebrahimi, S., Meier, F., Calandra, R., Darrell, T., Rohrbach, M.: Adversarial
  continual learning. The 2020 European Conference on Computer Vision (ECCV)
  (2020)

\bibitem{Edwards2017TowardsAN}
Edwards, H., Storkey, A.: Towards a neural statistician. ArXiv
  \textbf{abs/1606.02185} (2017)

\bibitem{finn17a}
Finn, C., Abbeel, P., Levine, S.: Model-agnostic meta-learning for fast
  adaptation of deep networks. International Conference on Machine Learning
  (2017)

\bibitem{finn2019online}
Finn, C., Rajeswaran, A., Kakade, S., Levine, S.: Online meta-learning. In:
  Proceedings of International Conference on Machine Learning (2019)

\bibitem{PMAML2018}
Finn, C., Xu, K., Levine, S.: Probabilistic model-agnostic meta-learning.
  Advances in Neural Information Processing Systems  (2018)

\bibitem{Genevay2018}
Genevay, A., Peyr{\'e}, G., Cuturi, M.: Learning generative models with
  sinkhorn divergences  (2018)

\bibitem{guo2020learning}
Guo, J., Zhu, X., Zhao, C., Cao, D., Lei, Z., Li, S.Z.: Learning meta face
  recognition in unseen domains. Conference on Computer Vision and Pattern
  Recognition (CVPR)  (2020)

\bibitem{guo2021metacorrection}
Guo, X., Yang, C., Li, B., Yuan, Y.: Metacorrection: Domain-aware meta loss
  correction for unsupervised domain adaptation in semantic segmentation.
  Conference on Computer Vision and Pattern Recognition (CVPR)  (2021)

\bibitem{vanhorn2018inaturalist}
Horn, G.V., Aodha, O.M., Song, Y., Cui, Y., Sun, C., Shepard, A., Adam, H.,
  Perona, P., Belongie, S.: The inaturalist species classification and
  detection dataset. IEEE Conference on Computer Vision and Pattern Recognition
  (CVPR)  (2018)

\bibitem{huang-etal-2020-semi}
Huang, X., Qi, J., Sun, Y., Zhang, R.: Semi-supervised dialogue policy learning
  via stochastic reward estimation. Proceedings of the 58th Annual Meeting of
  the Association for Computational Linguistics  (2020)

\bibitem{kingma2014adam}
Kingma, D.P., Ba, J.: Adam: A method for stochastic optimization. International
  Conference on Learning Representations  (2014)

\bibitem{EWC16}
Kirkpatrick, J., Pascanu, R., Rabinowitz, N., Veness, J., Desjardins, G., Rusu,
  A.A., Milan, K., Quan, J., Ramalho, T., Grabska-Barwinska, A., Hassabis, D.,
  Clopath, C., Kumaran, D., Hadsell, R.: Overcoming catastrophic forgetting in
  neural networks. Proceedings of the national academy of sciences  (2017)

\bibitem{Omniglot2011}
Lake, B., Salakhutdinov, R., Gross, J., Tenenbaum, J.: One shot learning of
  simple visual concepts. Conference of the Cognitive Science Society  (2011)

\bibitem{lee2019overcoming}
Lee, K., Lee, K., Shin, J., Lee, H.: Overcoming catastrophic forgetting with
  unlabeled data in the wild. 2019 IEEE/CVF International Conference on
  Computer Vision (ICCV)  (2019)

\bibitem{lee2019meta}
Lee, K., Maji, S., Ravichandran, A., Soatto, S.: Meta-learning with
  differentiable convex optimization. In: Proceedings of the IEEE Conference on
  Computer Vision and Pattern Recognition (CVPR) (2019)

\bibitem{li2019learning}
Li, X., Sun, Q., Liu, Y., Zheng, S., Zhou, Q., Chua, T.S., Schiele, B.:
  Learning to self-train for semi-supervised few-shot classification.
  Proceedings of the Advances in Neural Information Processing Systems  (2019)

\bibitem{liu2019learning}
Liu, Y., Lee, J., Park, M., Kim, S., Yang, E., Hwang, S.J., Yang, Y.: Learning
  to propagate labels: Transductive propagation network for few-shot learning.
  International Conference on Learning Representations  (2019)

\bibitem{lopezpaz2017gradient}
Lopez-Paz, D., Ranzato, M.: Gradient episodic memory for continual learning.
  Advances in Neural Information Processing Systems  (2017)

\bibitem{maji2013finegrained}
Maji, S., Rahtu, E., Kannala, J., Blaschko, M., Vedaldi, A.: Fine-grained
  visual classification of aircraft. https://arxiv.org/abs/1306.5151  (2013)

\bibitem{SNAILICLR18}
Mishra, N., Rohaninejad, M., Chen, X., Abbeel, P.: A simple neural attentive
  meta-learner. International Conference on Learning Representations  (2018)

\bibitem{munkhdalai2017meta}
Munkhdalai, T., Yu, H.: Meta networks. Proceedings of the 34th International
  Conference on Machine Learning  (2017)

\bibitem{nguyen2015deep}
Nguyen, A., Yosinski, J., Clune, J.: Deep neural networks are easily fooled:
  High confidence predictions for unrecognizable images. IEEE Conference on
  Computer Vision and Pattern Recognition (CVPR)  (2015)

\bibitem{nguyen2017variational}
Nguyen, C.V., Li, Y., Bui, T.D., Turner, R.E.: Variational continual learning.
  Proceedings of the International Conference on Learning Representations
  (2018)

\bibitem{Nilsback08}
Nilsback, M.E., Zisserman, A.: Automated flower classification over a large
  number of classes  (2008)

\bibitem{poole2019variational}
Poole, B., Ozair, S., van~den Oord, A., Alemi, A.A., Tucker, G.: On variational
  bounds of mutual information (2019)

\bibitem{raghu2020rapid}
Raghu, A., Raghu, M., Bengio, S., Vinyals, O.: Rapid learning or feature reuse?
  towards understanding the effectiveness of maml. International Conference on
  Learning Representations  (2020)

\bibitem{Ravi2017}
Ravi, S., Larochelle, H.: Optimization as a model for few-shot learning. In:
  International Conference on Learning Representations (2017)

\bibitem{ren2018metalearning}
Ren, M., Triantafillou, E., Ravi, S., Snell, J., Swersky, K., Tenenbaum, J.B.,
  Larochelle, H., Zemel, R.S.: Meta-learning for semi-supervised few-shot
  classification. International Conference on Learning Representations  (2018)

\bibitem{riemer2018learning}
Riemer, M., Cases, I., Ajemian, R., Liu, M., Rish, I., Tu, Y., Tesauro, G.:
  Learning to learn without forgetting by maximizing transfer and minimizing
  interference. International Conference on Learning Representations  (2019)

\bibitem{saha2021gradient}
Saha, G., Garg, I., Roy, K.: Gradient projection memory for continual learning.
  International Conference on Learning Representations  (2021)

\bibitem{pmlr-v48-santoro16}
Santoro, A., Bartunov, S., Botvinick, M., Wierstra, D., Lillicrap, T.:
  Meta-learning with memory-augmented neural networks. Proceedings of the 34th
  International Conference on Machine Learning  (2016)

\bibitem{metanet}
{Schmidhuber}, J.: A neural network that embeds its own meta-levels. IEEE
  International Conference on Neural Networks  (1993)

\bibitem{sehwag2021ssd}
Sehwag, V., Chiang, M., Mittal, P.: Ssd: A unified framework for
  self-supervised outlier detection. International Conference on Learning
  Representations  (2021)

\bibitem{smith2021memoryefficient}
Smith, J., Balloch, J., Hsu, Y.C., Kira, Z.: Memory-efficient semi-supervised
  continual learning: The world is its own replay buffer.
  https://arxiv.org/abs/2101.09536  (2021)

\bibitem{protonet17}
Snell, J., Swersky, K., Zemel, R.S.: Prototypical networks for few-shot
  learning. Advances in Neural Information Processing Systems  (2017)

\bibitem{tseng2020crossdomain}
Tseng, H.Y., Lee, H.Y., Huang, J.B., Yang, M.H.: Cross-domain few-shot
  classification via learned feature-wise transformation. Proceedings of the
  International Conference on Learning Representations  (2020)

\bibitem{matching16}
Vinyals, O., Blundell, C., Lillicrap, T., Kavukcuoglu, K., Wierstra, D.:
  Matching networks for one shot learning. https://arxiv.org/pdf/1606.04080.pdf
   (2016)

\bibitem{reservoirsampling}
Vitter, J.S.: Random sampling with a reservoir. Association for Computing
  Machinery (1985)

\bibitem{reservoir1985}
Vitter, J.S.: Random sampling with a reservoir. ACM Transactions on
  Mathematical Software  (1985)

\bibitem{vuorio2019multimodal}
Vuorio, R., Sun, S.H., Hu, H., Lim, J.J.: Multimodal model-agnostic
  meta-learning via task-aware modulation. Proceedings of the Advances in
  Neural Information Processing Systems  (2019)

\bibitem{wang2021ordisco}
Wang, L., Yang, K., Li, C., Hong, L., Li, Z., Zhu, J.: Ordisco: Effective and
  efficient usage of incremental unlabeled data for semi-supervised continual
  learning. IEEE Conference on Computer Vision and Pattern Recognition (CVPR)
  (2021)

\bibitem{Wang_2019_ICCV}
Wang, Y.X., Ramanan, D., Hebert, M.: Meta-learning to detect rare objects.
  Proceedings of the IEEE/CVF International Conference on Computer Vision
  (ICCV)  (October 2019)

\bibitem{wang2021meta}
Wang, Z., Duan, T., Fang, L., Suo, Q., Gao, M.: Meta learning on a sequence of
  imbalanced domains with difficulty awareness. In: Proceedings of the IEEE/CVF
  International Conference on Computer Vision. pp. 8947--8957 (2021)

\bibitem{Wang_2022_CVPR}
Wang, Z., Shen, L., Duan, T., Zhan, D., Fang, L., Gao, M.: Learning to learn
  and remember super long multi-domain task sequence. In: Proceedings of the
  IEEE/CVF Conference on Computer Vision and Pattern Recognition (CVPR) (2022)

\bibitem{pmlr-v162-wang22v}
Wang, Z., Shen, L., Fang, L., Suo, Q., Duan, T., Gao, M.: Improving task-free
  continual learning by distributionally robust memory evolution. In:
  International Conference on Machine Learning. pp. 22985--22998 (2022)

\bibitem{metafusion}
Wang, Z., Wang, X., Shen, L., Suo, Q., Song, K., Yu, D., Shen, Y., Gao, M.:
  Meta-learning without data via wasserstein distributionally-robust model
  fusion. In: The Conference on Uncertainty in Artificial Intelligence (UAI),
  2022 (2022)

\bibitem{Wang2020Bayesian}
Wang, Z., Zhao, Y., Yu, P., Zhang, R., Chen, C.: Bayesian meta sampling for
  fast uncertainty adaptation. International Conference on Learning
  Representations  (2020)

\bibitem{WelinderEtal2010}
Welinder, P., Branson, S., Mita, T., Wah, C., Schroff, F., Belongie, S.,
  Perona, P.: {Caltech-UCSD Birds 200}  (2010)

\bibitem{yoon2018lifelong}
Yoon, J., Yang, E., Lee, J., Hwang, S.J.: Lifelong learning with dynamically
  expandable networks. International Conference on Learning Representations
  (2018)

\bibitem{metakernel}
Zhou, Y., Wang, Z., Xian, J., Chen, C., Xu, J.: Meta-learning with neural
  tangent kernels. International Conference on Learning Representations  (2021)

\bibitem{zhuang2018wildfish}
Zhuang, P., Wang, Y., Qiao, Y.: Wildfish: A large benchmark for fish
  recognition in the wild  (2018)

\end{thebibliography}
\clearpage
\appendix

\section{Dataset Details} \label{appendix:dataset}

\paragraph{Miniimagenet \cite{matching16}} A dataset selected from ImageNet with 100 different classes, each with 600 images. All images are the same size of 84×84. We adopt the same splits of \cite{Ravi2017} with meta train/validation/test splits being 64/16/20 classes. 

\paragraph{CIFARFS \cite{bertinetto2019metalearning}} A dataset randomly sampled from CIFAR-100. The meta train/validation/test splits are of 64/16/20 classes respectively. We follow the split of \cite{bertinetto2019metalearning}. 

\paragraph{Omniglot \cite{Omniglot2011}} An image dataset of 1623 handwritten characters from 50 different alphabets, with 20 examples per class. We follow the setup and data split in \cite{matching16} and use this dataset as OOD data.

\paragraph{CUB \cite{WelinderEtal2010}} A dataset for fine-grained classification on 200 different bird species. The meta train/validation/test splits are of 100/50/50 classes respectively. We follow the same split of \cite{chen2019closerfewshot}.

\paragraph{AIRCRAFT \cite{maji2013finegrained}} A dataset of images for aircrafts model classification consisting of 102 categories, with 100 images per class. The dataset is split into 70/15/15 classes for meta- training/validation/test. We follow the split in \cite{vuorio2019multimodal}.

\paragraph{Plantae \cite{vanhorn2018inaturalist}}  A large dataset consisting of approximately 100K images with 200 randomly selected plant species. We follow the split of \cite{tseng2020crossdomain} and split the dataset into 100/50/50 classes for meta- training/validation/test.

\paragraph{Butterfly \cite{chen2018finegrained}} A large scale fine-grained dataset of Butterfly, they have collected 25279 butterfly
images from 200 species, with each species containing at least 30 images. We only keep the classes with more than 50 images and got 185 classes. We randomly split the dataset into 100/40/45 classes for meta-training/validation/test.

\paragraph{Vggflower \cite{Nilsback08}} there are 102 flower categories and contains between 40 and 258 images for
each class. We use this dataset for OOD data.

\paragraph{Fish \cite{zhuang2018wildfish}}
A large scale dataset of fish species. It consists of 1000 fish categories with a total of 54,459 images, we use this dataset for OOD data.

\section{Implementation details} \label{app:implementdetails}

\textbf{Memory buffer update}

The memory is updated by online reservoir sampling (RS) \cite{reservoir1985}, at each time step $t$,  RS operates on task unit level. RS ensures each task has the same probability to be stored in the memory buffer without knowing the total number of tasks in advance. The algorithm works by maintaining a reservoir of size $k$ tasks. As long as the memory buffer is not full, each new incoming task will be stored in the memory buffer. When the memory buffer is full, at time step $i$, when the task $i$ arrives, the algorithm then generates a random number $j$ between (and including) 1 and $i$. If $j$ is at most $k$, then the task is selected and replaces one task in the reservoir. Otherwise, the task $i$ is discarded. The memory buffer is updated online and real time. Whether each task is determined to be stored in the memory or not depends on whether the random number $j$ is less than k or not.

\textbf{Implementation details.} 
 We use a five-layer CNN with 64 filters of kernel size being 3 for meta-learning. Similar architecture is commonly used in existing meta-learning literature. We do not use any pre-trained network feature extractors, which assume all the training tasks data are available before training, and this violates our problem setting that future incoming tasks are completely unknown. Methods are evaluated on the proposed benchmark described in Section \ref{sec:problemsetting}. We perform experiments on different dataset ordering, with the default ordering being  Plantae, CUB, MiniImagenet, CIFARFS, Aircraft, Butterfly. For each dataset, we randomly sample 12K tasks, we thus have 72K tasks in total. This task sequence is much larger than existing continual learning models considered. The evolving episodes are constructed and described in Section \ref{sec:problemsetting}.  All experiments are averaged over five independent runs. We create an additional split for each dataset to separate the images of each class into disjoint
labeled and unlabeled sets. Following \cite{ren2018metalearning}, we sample 40\% of the images of
each class to form the labeled split, and the remaining 60\% can only be used in the unlabeled portion
of episodes. We randomly sample 10 data points from the unlabeled portion of each chosen class to form the unlabeled ID data $\mathcal{U}_{id}$ for each episode.  The unlabeled OOD data $\mathcal{U}_{ood}$ for each episode is produced by sampling $R$ unlabeled data from the mixture OOD dataset of \textit{Omniglot} \cite{Omniglot2011}, \textit{Vggflower} \cite{Nilsback08} and \textit{Fish} \cite{zhuang2018wildfish}. $R=50$ by default. This way of constructing multimodal OOD data is more challenging and more realistic than existing SSFSL works, which sample OOD data from the same meta training dataset. More implementation details are given in Appendix \ref{app:implementdetails}. We provide code in supplementary materials.

We randomly sample 12000 tasks from each dataset. We sequentially train on each dataset for 6000 iterations in the dataset sequence. At each iteration, we randomly sample 2 tasks (meta batch size) from current dataset.  The hyperparameter is determined by grid search, with $\lambda = \{1e-4, 1e-5, 1e-6\}$ and $\beta = \{0.01, 0.005, 0.003, 0.001\}$. The grid search results for 5-way 5-shot learning are in Table \ref{tab:hyper}. The adopted values are $\lambda = 1e-5$ and $\beta=0.003$.
The OOD sampling is performed by uniformly sampling equal number of OOD data from each OOD dataset. We use Adam optimizer \cite{kingma2014adam} to optimize the model with learning rate of 1e-3. For theorem 3, we set $a(y) = 1$ for simplicity. The function in $f(\eb_{id}, \pb_{c^{\prime}})$ is one layer network with 10 hidden units.

\section{More experimental results} \label{sec:order}

\textbf{Effect of domain ordering}

Order 2 :   Butterfly,  CUB,  CIFARFS, Plantae, MiniImagenet, Aircraft. 
5-way, 1-shot and 5-shot learning results are shown in Table \ref{tab:metabaseline2} and Table \ref{tab:continual2}.

Order 3 :   CUB, CIFARFS, Plantae, MiniImagenet, Butterfly, Aircraft. Results are shown in Table \ref{tab:metabaseline3} and Table \ref{tab:continual3}

\begin{table}[H]
\centering
\captionsetup{width=0.85\linewidth}
\caption{5-way, 1-shot and 5-shot classification accuracy comparing to meta-learning baselines with domain order 2. Top rows are meta learning \textit{without unlabeled data}, bottom rows are meta learning methods \textit{with unlabeled data}.}
\begin{adjustbox}{scale=0.85,tabular= lccccccc,center}
\begin{tabular}{lrrrrrrrrr}
\toprule
Algorithm & 1-Shot &  5-Shot &\\
\midrule  
ProtoNet  &$27.91\pm 0.95$  &  $39.12\pm 0.76$ & \\
\midrule  
ANIL   &$27.25\pm 0.82$ & $38.58\pm 0.65$  & \\
\midrule
\midrule
MSKM  &$30.15\pm 0.89$  & $41.02\pm 0.75$ \\
\midrule
LST & $32.51\pm 0.71$ &  $41.78\pm 0.85$ &  \\
\midrule
TPN  &$31.81\pm 0.72$  &  $42.81\pm 0.62$  \\
\midrule
Ours  &\bm{$42.49\pm 0.57$} & \bm{$55.37\pm 0.71$}   \\
\bottomrule
\end{tabular}
\label{tab:metabaseline2}
\end{adjustbox}
\end{table}

\begin{table}[H]
\centering
\captionsetup{width=0.85\linewidth}
\caption{5-way, 1-shot and 5-shot classification accuracy compared to continual learning baselines.}
\begin{adjustbox}{scale=0.85,tabular= lccccccc,center}
\begin{tabular}{lrrrrrrrrr}
\toprule
Algorithm & 1-Shot &  5-Shot &\\
\midrule  
Semi-ER  & $39.05\pm 0.91$ & $51.28\pm 0.78$  \\
\midrule
Semi-AGEM  & $39.48\pm 0.97$ & $51.51\pm 0.83$ \\
\midrule
Semi-MER  & $39.65\pm 0.82$& $51.42\pm 0.68$ & \\
\midrule
Semi-GPM & $38.89\pm 0.85$&  $51.03\pm 0.71$ \\
\midrule
Semi-DEGCL& $39.81\pm 0.73$&  $51.68\pm 0.62$ &  \\
\midrule
Ours   & \bm{$42.49\pm 0.57$} & \bm{$55.37\pm 0.71$} \\
\midrule
Joint-training & $49.91\pm 0.79$ &  $61.78\pm 0.75$\\
\midrule
\bottomrule
\end{tabular}
\label{tab:continual2}
\end{adjustbox}
\end{table}

\begin{table}[H]
\centering
\captionsetup{width=0.85\linewidth}
\caption{5-way, 1-shot and 5-shot classification accuracy comparing to meta-learning baselines with domain order 2. Top rows are meta learning \textit{without unlabeled data}, bottom rows are meta learning methods \textit{with unlabeled data}.}
\begin{adjustbox}{scale=0.85,tabular= lccccccc,center}
\begin{tabular}{lrrrrrrrrr}
\toprule
Algorithm & 1-Shot &  5-Shot &\\
\midrule  
ProtoNet  &$28.55\pm 1.03$ & $40.41\pm 0.97$ \\
\midrule  
ANIL  &$28.24\pm 1.15$ &  $39.97\pm 0.83$ & \\
\midrule
\midrule
MSKM &  $31.17\pm 0.96$ & $41.46\pm 0.69$  \\
\midrule
LST & $31.61\pm 0.81$ & $41.31\pm 0.80$ \\
\midrule
TPN  & $32.80\pm 0.71$ & $42.83\pm 0.71$  \\
\midrule
Ours   & \bm{$43.36\pm 0.61$} &  \bm{$56.89\pm 0.58$} \\
\bottomrule
\end{tabular}
\label{tab:metabaseline3}
\end{adjustbox}
\end{table}

\begin{table}[H]
\centering
\captionsetup{width=0.85\linewidth}
\caption{5-way, 1-shot and 5-shot classification accuracy compared to continual learning baselines.}
\begin{adjustbox}{scale=0.85,tabular= lccccccc,center}
\begin{tabular}{lrrrrrrrrr}
\toprule
Algorithm & 1-Shot &  5-Shot &\\
\midrule  
Semi-ER  & $38.87\pm 0.76$ &  $52.75\pm 0.82$ \\
\midrule
Semi-AGEM  &$39.29\pm 0.73$ & $52.97\pm 0.94$ \\
\midrule
Semi-MER  &$39.41\pm 0.68$ & $53.08\pm 0.87$\\
\midrule
Semi-GPM &$39.09\pm 0.81$ &  $52.83\pm 0.76$ &  \\
\midrule
Semi-DEGCL & $39.46\pm 0.77$ & $53.37\pm 0.73$  &  \\
\midrule
Ours    & \bm{$43.36\pm 0.61$} &  \bm{$56.89\pm 0.58$} \\
\midrule
Joint-training & $49.91\pm 0.79$ &  $61.78\pm 0.75$\\
\midrule
\bottomrule
\end{tabular}
\label{tab:continual3}
\end{adjustbox}
\end{table}

\begin{table*}
\centering 
\caption{Sensitivity analysis on hyper parameters} 

\begin{tabular}{l|*{5}{c}}\hline 
\backslashbox{$\lambda$}{$\beta$}
&\makebox[3em]{0.001}&\makebox[3em]{0.003}&\makebox[3em]{0.005}
&\makebox[3em]{0.01}\\\hline
1e-4 &$52.86\pm 0.79$ &$53.28\pm 0.75$ &$53.49\pm 0.67$ &  $52.65\pm 0.79$   \\\hline
1e-5 & $52.61\pm  0.85$&$53.96\pm 0.71$ &$53.60\pm 0.82$ &$52.73\pm 0.93$  \\\hline
1e-6 &$51.85\pm 0.58$ &$52.06\pm 0.49$ &$52.15\pm 0.70$&$52.03\pm 0.67$ \\\hline
\end{tabular}
\label{tab:hyper}
\end{table*}

\textbf{Sensitivity analysis on hyper parameters}

We performed sensitivity analysis on hyperparameters, the result is summarized in Table \ref{tab:hyper}. $\beta$ controls the magnitude of optimal transport regularization, the result indicate the model performance is positively correlated with $\beta$ when it increases from 0.1 until optimal tradeoff is reached, after which performance deteriorates with over regularization when $\beta$ reaches 1.5. Similar trend is observed in mutual information regularization with $\lambda$. 

\textbf{Effect of memory size}

Table \ref{tab:memorysize} shows the effect of different memory size. With memory size increases, the model performance also increases accordingly. 

\begin{table}[H]
\centering
\captionsetup{width=0.85\linewidth}
\caption{5-way, 1-shot and 5-shot classification accuracy with different memory size.}
\begin{adjustbox}{scale=0.85,tabular= lccccccc,center}
\begin{tabular}{lrrrrrrrrr}
\toprule
Memory size & 1-Shot &  5-Shot &\\
\midrule
50  &$40.11\pm 0.73$ & $52.75\pm 0.77$   \\
\midrule
100 &$40.79\pm 0.58$ & $53.28\pm 0.65$ \\
\midrule
200  &$41.25\pm 0.64$ & $53.96\pm 0.71$ \\
\midrule
\bottomrule
\end{tabular}
\label{tab:memorysize}
\end{adjustbox}
\end{table}

\textbf{Ablation Study results}

\begin{minipage}{0.5\linewidth}
\small
\begin{figure}[H] 
     \centering
     \begin{subfigure}[b]{1\textwidth}
         \centering
         \includegraphics[width=\textwidth]{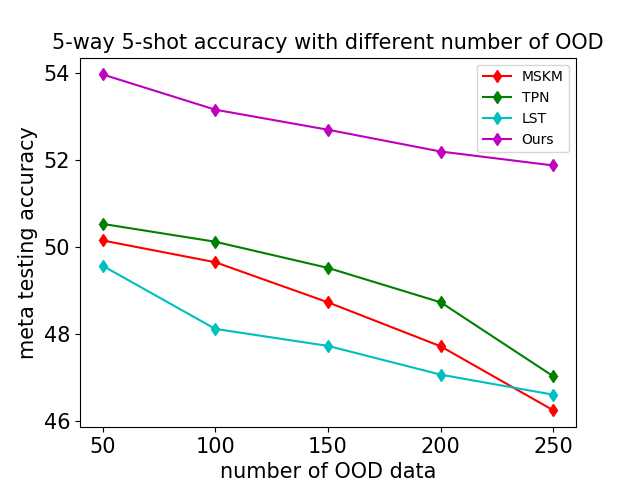}
     \end{subfigure}
     \caption{Sensitivity of baselines and our method to OOD data}
    \label{fig:ood}
\end{figure}
\end{minipage}

\begin{table}[H]
\centering
\captionsetup{width=0.9\linewidth}
\caption{Running time (seconds). }
\begin{adjustbox}{scale=0.85,tabular= lccccccc,center}
\begin{tabular}{lrrrrrrrrr}
\toprule
Algorithm &{5-Shot (250 OOD)} & {5-Shot (50 OOD)} \\
\midrule
MSKM &\textbf{85} & \textbf{43}  \\
TPN &149 & 86 \\
LST &137 & 79 \\
\midrule
ORDER (Ours) &122 & 71\\
\bottomrule
\end{tabular}
\label{eq:runtime}
\end{adjustbox}
\end{table}

\textbf{Efficiency Evaluation}
To investigate the computational efficiency, we report the time cost of baselines and our method running on 200 iterations. The results are reported in Table \ref{eq:runtime}. Our method is more efficient than TPN and LST, but slower than MSKM.

\begin{table}[H]
\centering
\captionsetup{width=0.85\linewidth}
  \caption{\small Ablation study of model components.}
  \begin{tabular}{cccccccccccc}
  \toprule  
     \centering
MI  &OT   & ACC (5 shot 50 OOD) & ACC (5 shot 250 OOD)\\ \midrule
\xmark &\xmark & $50.35\pm 0.76$   & $46.25\pm 0.91$ \\
 \xmark &\checkmark & $51.18\pm 0.82$  & $47.69\pm 0.87$ \\
 \xmark &\checkmark ($\mathcal{U}_{id}$) & $51.72\pm 0.61$  & $48.57\pm 0.80$ \\
\checkmark &\xmark& $52.63\pm 0.77$ &  $50.23\pm 0.62$ \\
\checkmark  &\checkmark ($\mathcal{U}_{id}$) & $53.96\pm 0.71$  & $51.87\pm 0.68$\\
    \bottomrule
  \end{tabular}
\label{tab:ablation}
\end{table}

\begin{table}[H]
\centering
\captionsetup{width=0.9\linewidth}
\centering
  \caption{\small Fine-grained ablation study analysis of using different parts of unlabeled data and mutual information regularizer.}
\begin{adjustbox}{scale=0.85,tabular= lccccccc,center}
\begin{tabular}{cccccccccccc}
  \toprule  
     \centering
$\mathcal{U}_{id}$ & $\mathcal{U}_{ood}$ & $\mathcal{I}_{id}$  & $\mathcal{I}_{ood}$  & ACC (5 shot 50 OOD) & ACC (5 shot 250 OOD)\\ \midrule
\checkmark & \checkmark &\xmark &\xmark & $50.35\pm 0.76$   & $46.25\pm 0.91$ \\
\checkmark & \xmark & \xmark &\xmark  & $51.67\pm 0.85$   & $48.36\pm 0.82$  \\
\checkmark &\xmark& \checkmark &\xmark & $52.87\pm 0.89$   & $49.98\pm 0.70$  \\
\checkmark  &\checkmark& \checkmark  &\checkmark& $53.96\pm 0.71$  & $51.87\pm 0.68$\\
    \bottomrule
  \end{tabular}
  \end{adjustbox}
  \label{tab:ablation2}
\end{table}

\section{Theorem Proof} \label{app:theory}

\textbf{Theorem 1} For a task $\mathcal{T} = \{\mathcal{S}, \mathcal{U}, \mathcal{Q}\} $, suppose the unlabeled OOD data embedding $\eb_{ood} = h_{\thetab}(\xb), \xb\in \mathcal{U}_{ood}$. The nearest prototype corresponds to $\eb_{ood}$ is $\pb_{c^{\prime}}$, where $c^{\prime} =  \argmin_{c}  ||\eb_{ood} - \pb_c||$. Given a collection of samples $\{(\eb_{ood}^{i}, \pb_{c^{\prime}}^{i})\}_{i=1}^{i=L} \sim P(\eb_{ood}, \pb_{c^{\prime}})$, the  variational upper bound of mutual information $ \mathcal{I} (\eb_{ood}, \pb_{c^{\prime}})$ is 
 \begin{align}
 \small
      &\mathcal{I} (\eb_{ood}, \pb_{c^{\prime}}) \nonumber \\
      &\leq \sum_{i=1}^{i=L} \log P(\pb_{c^{\prime}}^i|\eb_{ood}^i) -  \sum_{i=1}^{i=L}\sum_{j=1}^{j=L} \log P(\pb_{c^{\prime}}^i|\eb_{ood}^j) \nonumber \\
     & = \mathcal{I}_{ood}. 
 \end{align}
 The bound is tight (equality holds) when $\pb_{c^{\prime}}$ and $\eb_{ood}$ are independent.

\textbf{Proof}
\begin{align*}
    &\quad \mathbb{E}_{P(\eb_{ood}, \pb_{c^{\prime}})} [\log P(\pb_{c^{\prime}}|\eb_{ood})]\\
    &\quad -\mathbb{E}_{P(\pb_{c^{\prime}})P(\eb_{ood})}[\log P(\pb_{c^{\prime}}|\eb_{ood})] - \mathcal{I} (\eb_{ood}, \pb_{c^{\prime}}) \\
    &= 
    \mathbb{E}_{P(\eb_{ood}, \pb_{c^{\prime}})} [\log P(\pb_{c^{\prime}}|\eb_{ood})]\\
    & \quad -\mathbb{E}_{P(\pb_{c^{\prime}})P(\eb_{ood})}[\log P(\pb_{c^{\prime}}|\eb_{ood})] \\
    &\quad - \mathbb{E}_{P(\eb_{ood}, \pb_{c^{\prime}})} [\log P(\pb_{c^{\prime}}|\eb_{ood}) - \log P(\pb_{c^{\prime}})] \\
    &=  
    \mathbb{E}_{P(\eb_{ood}, \pb_{c^{\prime}})} [\log P(\pb_{c^{\prime}})] -\mathbb{E}_{P(\pb_{c^{\prime}})P(\eb_{ood})}[\log P(\pb_{c^{\prime}}|\eb_{ood})] \\
    &=
    \mathbb{E}_{P( \pb_{c^{\prime}})} [\log P(\pb_{c^{\prime}}) -\mathbb{E}_{P(\eb_{ood})}[\log P(\pb_{c^{\prime}}|\eb_{ood})]]
\end{align*}

Following \cite{pmlrv119cheng20b}, 

\begin{equation}
    P(\pb_{c^{\prime}}) = \mathbb{E}_{P(\eb_{ood})} [P( \pb_{c^{\prime}}|\eb_{ood})]
\end{equation}

By Jensen's inequality,
\begin{equation*}
\log \mathbb{E}_{P(\eb_{ood})} [P( \pb_{c^{\prime}}|\eb_{ood})] \geq \mathbb{E}_{P(\eb_{ood})}[\log P(\pb_{c^{\prime}}|\eb_{ood})]
\end{equation*}

We then get the following bounds:
\begin{align}
    \mathcal{I} (\eb_{ood}, \pb_{c^{\prime}}) 
    &\leq \mathbb{E}_{P(\eb_{ood}, \pb_{c^{\prime}})} [\log P(\pb_{c^{\prime}}|\eb_{ood})] \\ &-\mathbb{E}_{P(\pb_{c^{\prime}})P(\eb_{ood})}[\log P(\pb_{c^{\prime}}|\eb_{ood})]\nonumber
\end{align}

 Given a collection of samples $\{(\eb_{ood}^{i}, \pb_{c^{\prime}}^{i})\}_{i=1}^{i=L} \sim P(\eb_{ood}, \pb_{c^{\prime}})$, the upper bound becomes: 

 \begin{align}\label{eq:ood2}
      \mathcal{I} (\eb_{ood}, \pb_{c^{\prime}}) 
      &\leq \sum_{\pb_{c^{\prime}}, \eb_{ood}} \log P(\pb_{c^{\prime}}|\eb_{ood}) \\
      &-  \sum_{c=1}^{c=N}\sum_{j=1}^{j=M} \log P(\pb_{c^{\prime}}|\eb_{ood}^j) = \mathcal{I}_{ood}. \nonumber
 \end{align}

\textbf{Lemma 2} 
 For a task $\mathcal{T} = \{\mathcal{S}, \mathcal{U}, \mathcal{Q}\} $, suppose the  unlabeled ID data embedding $\eb_{id} = h_{\thetab}(\xb), \xb\in \mathcal{U}_{id}$. The nearest prototype corresponds to $\eb_{id}$ is $\pb_{c^{\prime}}$, where $c^{\prime} =  \argmin_{c}  ||\eb_{id} - \pb_c||$. Given a collection of samples $\{(\eb_{id}^{i}, \pb_{c^{\prime}}^{i})\}_{i=1}^{i=L} \sim P(\eb_{id}, \pb_{c^{\prime}})$, the variational lower bound of $\mathcal{I} (\eb_{id}, \pb_{c^{\prime}})$ is :
 \begin{equation}
 \small
       \mathcal{I} (\eb_{id}, \pb_{c^{\prime}}) \geqslant \sum_{i=1}^{i=L} f(\eb_{id}^i, \pb_{c^{\prime}}^i) - \sum_{j=1}^{j=L}\log \sum_{i=1}^{i=L} e^{f(\eb_{id}^{i}, \pb_{c^{\prime}}^j)}.
 \end{equation}
 This bound is tight if $f(\eb_{id}, \pb_{c^{\prime}}) = \log P(\pb_{c^{\prime}}|\eb_{id}) + c(\pb_{c^{\prime}})$.

\textbf{Proof}

With $q(\eb_{id} |\pb_{c^{\prime}} )$ to approximate $P(\eb_{id} |\pb_{c^{\prime}} )$, we can get the following inequality (following the argument in Barber-Agakov lower bound \cite{VIbound}):

\begin{align*}
    \mathcal{I} (\eb_{id}, \pb_{c^{\prime}}) 
    &= \mathbb{E}_{P(\eb_{id},      \pb_{c^{\prime}})} [\frac{q(\eb_{id} |\pb_{c^{\prime}} )} {P(\eb_{id})}] \\
    & \quad + \mathbb{E}_{P(\eb_{id})}[\mathbb{KL}(P(\eb_{id}| \pb_{c^{\prime}}) |q(\eb_{id}|\pb_{c^{\prime}}))] \\
    & \geq \mathbb{E}_{P(\eb_{id},      \pb_{c^{\prime}})} [q( \eb_{id}|\pb_{c^{\prime}}))] + H(\eb_{id})
\end{align*}

$H(\eb_{id})$ is the entropy of the variable $\eb_{id}$. If $q(\eb_{id}|\pb_{c^{\prime}}))$ is specified as: 

\begin{equation}
    q(\eb_{id}|\pb_{c^{\prime}}) = \frac{P(\eb_{id})}{Z(\pb_{c^{\prime}})} e^{f( \eb_{id}, \pb_{c^{\prime}})}
\end{equation}

where $f( \eb_{id}, \pb_{c^{\prime}})$ is a value function parametrized by a neural network. $Z(\pb_{c^{\prime}})$ is a partition function and is specified as following:

\begin{equation}
    Z(\pb_{c^{\prime}}) = \mathbb{E}_{P(\eb_{id})}[e^{f(\eb_{id}, \pb_{c^{\prime}})}]
\end{equation}

Then, the above lower bound becomes

\begin{equation}\label{eq:newbound}
    \mathbb{E}_{P(\eb_{id},      \pb_{c^{\prime}})} [f(\eb_{id}, \pb_{c^{\prime}}))] -  \mathbb{E}_{P(\pb_{c^{\prime}})} [\log Z(\pb_{c^{\prime}}))]
\end{equation}

Given a collection of samples $\{(\eb_{id}^{i}, \pb_{c^{\prime}}^{i})\}_{i=1}^{i=L} \sim P(\eb_{id}, \pb_{c^{\prime}})$, the bounds become

 \begin{equation}
 \small
       \mathcal{I} (\eb_{id}, \pb_{c^{\prime}}) \geqslant \sum_{i=1}^{i=L} f(\eb_{id}^i, \pb_{c^{\prime}}^i) - \sum_{j=1}^{j=L}\log \sum_{i=1}^{i=L} e^{f(\eb_{id}^{i}, \pb_{c^{\prime}}^j)}.
 \end{equation}

Based on Lemma \ref{eq:lemma}, we derive a tractable lower bound as Theorem \ref{theorem:lower}.

\textbf{Theorem 3}
 For a task $\mathcal{T} = \{\mathcal{S}, \mathcal{U}, \mathcal{Q}\} $, suppose the unlabeled ID data embedding $\eb_{id} = h_{\thetab}(\xb), \xb\in \mathcal{U}_{id}$. The nearest prototype corresponds to $\eb_{id}$ is $\pb_{c^{\prime}}$, where $c^{\prime} =  \argmin_{c}  ||\eb_{id} - \pb_c||$. Given a collection of samples $\{(\eb_{id}^{i}, \pb_{c^{\prime}}^{i})\}_{i=1}^{i=L} \sim P(\eb_{id}, \pb_{c^{\prime}})$, the variational lower bound of $\mathcal{I} (\eb_{id}, \pb_{c^{\prime}})$ is:
 \begin{align*}
 \small
     \mathcal{I} (\eb_{id}, \pb_{c^{\prime}}) \geqslant
     & -  \sum_{j=1}^{j=L} \left[\frac{\sum_{i=1}^{i=L}  e^{f(\eb_{id}^i, \pb_{c^{\prime}}^j)} }{a(\pb_{c^{\prime}}^j)} + \log(a(\pb_{c^{\prime}}^j))-1\right] \\
     &+ \sum_{i=1}^{i=L} f(\eb_{id}^i, \pb_{c^{\prime}}^i)  = \mathcal{I}_{id}. 
 \end{align*}
 The bound is tight (equality holds) when $f(\eb_{id}, \pb_{c^{\prime}}) = \log P(\pb_{c^{\prime}}|\eb_{id}) + c(\pb_{c^{\prime}})$ and $a(\pb_{c^{\prime}}) = \mathbb{E}_{p(\eb_{id})} e^{f(\eb_{id}, \pb_{c^{\prime}})}$. $a(\pb_{c^{\prime}})$ is any function that  $a(\pb_{c^{\prime}})>0$.

\textbf{Proof}

Define the function $\delta(x, y) = \log (x) - \frac{x}{y} - \log (y) + 1$, where $x>0, y>0$

Take the derivative $\frac{\partial \delta(x, y)}{\partial x} = \frac{y-x}{xy}$, when $x<y$, $\frac{\partial \delta(x, y)}{\partial x}>0$; when $x>y$, $\frac{\partial \delta(x, y)}{\partial x}<0$. This implies that when $x= y$, $\delta(x, y)$ achieves maximum. Thus, $\delta(x, y) \leq \delta(y, y) = 0$. We can obtain the following inequality \cite{poole2019variational}

\begin{equation}\label{eq:ineqaulity1}
 \log (x) \leq \frac{x}{y} + \log (y) - 1
\end{equation}

By the inequality of \ref{eq:ineqaulity1}, we can get the following inequality:
\begin{equation} \label{eq:inequal}
    \log Z(\pb_{c^{\prime}})\leq \frac{Z(\pb_{c^{\prime}})}{a(\pb_{c^{\prime}})} + \log (a(\pb_{c^{\prime}})) - 1
\end{equation}

Follows from the above lemma, and plug the Equation \ref{eq:inequal} into Equation \ref{eq:newbound}, this theorem follows 

 \begin{align}\label{eq:id2}
     \mathcal{I} (\eb_{id}, \pb_{c^{\prime}}) &\geqslant \mathbb{E}_{p(\eb_{id}, \pb_{c^{\prime}})} [f(\eb_{id}, \pb_{c^{\prime}})] \\
     &\quad - \mathbb{E}_{p(\pb_{c^{\prime}})} [\frac{\mathbb{E}_{p(\eb_{id})} e^{f(\eb_{id}, \pb_{c^{\prime}})} }{a(\pb_{c^{\prime}})} \!+\! \log(a(\pb_{c^{\prime}}))\!-\!1] \nonumber\\
     &= \mathcal{I}_{id}.  \nonumber
 \end{align}
 
 Given a collection of samples $\{(\eb_{id}^{i}, \pb_{c^{\prime}}^{i})\}_{i=1}^{i=L} \sim P(\eb_{id}, \pb_{c^{\prime}})$, the bounds can be obtained in the following: 
  \begin{align*}
 \small
     \mathcal{I} (\eb_{id}, \pb_{c^{\prime}}) &\geqslant  -  \sum_{j=1}^{j=L} \left[\frac{\sum_{i=1}^{i=L}  e^{f(\eb_{id}^i, \pb_{c^{\prime}}^j)} }{a(\pb_{c^{\prime}}^j)} + \log(a(\pb_{c^{\prime}}^j))-1\right] \\
     &\quad + \sum_{i=1}^{i=L} f(\eb_{id}^i, \pb_{c^{\prime}}^i)   = \mathcal{I}_{id}. 
 \end{align*}
 Then, we obtain the desired bound.

\end{document}